\documentclass[11pt]{article}

\usepackage[final]{acl}

\usepackage{times}
\usepackage{latexsym}

\usepackage[T1]{fontenc}

\usepackage[utf8]{inputenc}

\usepackage{microtype}

\usepackage{inconsolata}

\usepackage{graphicx}

\usepackage{graphicx}
\usepackage{booktabs} 
\usepackage{multirow}
\usepackage{adjustbox}
\usepackage{float} 
\usepackage{colortbl}
\usepackage{xcolor}
\usepackage{longtable}
\usepackage{arydshln}
\usepackage{pifont}
\usepackage{array}
\usepackage[utf8]{inputenc}
\usepackage{subcaption}
\usepackage{algorithm}
\usepackage{algorithmic}
\usepackage{listings}
\usepackage{wrapfig}  
\usepackage{enumitem}
\usepackage{arydshln}
\usepackage{hyperref} 
\usepackage{stfloats}

\usepackage{tikz}
\usepackage{listings}

\usepackage{amsmath}
\usepackage{amssymb}
\usepackage{appendix}
\usepackage{placeins}

\usepackage[most]{tcolorbox}

\title{PaTaRM: Bridging Pairwise and Pointwise Signals via Preference-Aware Task-Adaptive Reward Modeling}

\author{
	Ai Jian$^{1,\dagger}$, Jingqing Ruan$^{2,\ddagger}$, Xing Ma$^{2}$, Xiaoyun Zhang$^{3}$, Dailin Li$^{2}$ \\
	\bfseries Weipeng Zhang$^{2}$, Ke Zeng$^{2}$, Xunliang Cai$^{2}$ \\
	\normalfont $^{1}$Beijing University of Posts and Telecommunications, Beijing, China \\
	$^{2}$Meituan, Beijing, China \\
	$^{3}$University of Chinese Academy of Sciences \\
	\texttt{jianai@bupt.edu.cn},
	\texttt{ruanjingqing@meituan.com}
}

\newcommand{\ourtitle}{PaTaRM}

\begin{document}
\maketitle
\begingroup
\renewcommand{\thefootnote}{\fnsymbol{footnote}}
\footnotetext[2]{Work done during an internship at Meituan.}
\footnotetext[3]{Corresponding author.}
\endgroup

\begin{abstract}
	Reward models (RMs) are central to reinforcement learning from human feedback (RLHF), providing the critical supervision signals that align large language models (LLMs) with human preferences. 
	Generative reward models (GRMs) provide greater interpretability than traditional scalar RMs, but they come with a critical trade-off: pairwise methods are hindered by a training-inference mismatch, while pointwise methods require expensive absolute annotations.
	To bridge this gap, we propose the \textbf{P}reference-\textbf{a}ware \textbf{T}ask-\textbf{a}daptive \textbf{R}eward \textbf{M}odel (\textbf{\ourtitle}).
	Unlike prior approaches, \ourtitle~enables robust pointwise training using readily available pairwise data via a novel \textit{Preference-Aware Reward (PAR)} mechanism, eliminating the need for explicit rating labels. Furthermore, it incorporates a \textit{task-adaptive rubric} system that dynamically generates instance-specific criteria for precise evaluation. Extensive experiments demonstrate that \ourtitle~achieves an average relative improvement of \textbf{8.7\%} over the corresponding base models on RewardBench and RMBench across the Qwen3-8B and Qwen3-14B backbones.
	Crucially, when used as a reward model for downstream RLHF, it yields an average relative improvement of \textbf{13.6\%} over the corresponding base policies on IFEval and InfoBench, validating its effectiveness for policy alignment.
	Our code, data, and checkpoints are available at \url{https://huggingface.co/AIJian/PaTaRM}.
\end{abstract}

\section{Introduction}

Reward models (RMs) are fundamental to reinforcement learning from human feedback (RLHF), serving as the critical supervision signals that guide large language models (LLMs) toward human-aligned behaviors. 
The predominant approach trains scalar reward models as discriminative classifiers that assign numerical scores to candidate responses, typically through the Bradley-Terry model~\citep{skywork,internlm2,eurus,BT}. 
While effective for basic preference alignment, scalar RMs exhibit significant limitations: they fail to fully leverage the generative and reasoning capabilities of LLMs~\citep{rmr1}, often capturing superficial correlations rather than genuine human preferences~\citep{nexttoken}. Moreover, they are prone to overfitting and sensitive to distribution shifts~\citep{sync}. 
To address these limitations, generative reward models (GRMs) have emerged as a promising alternative, offering more structured and interpretable evaluations of model outputs~\citep{rrm,rewardanything}.
\begin{figure}[t]
	\centering
	\includegraphics[width=1.0\columnwidth]{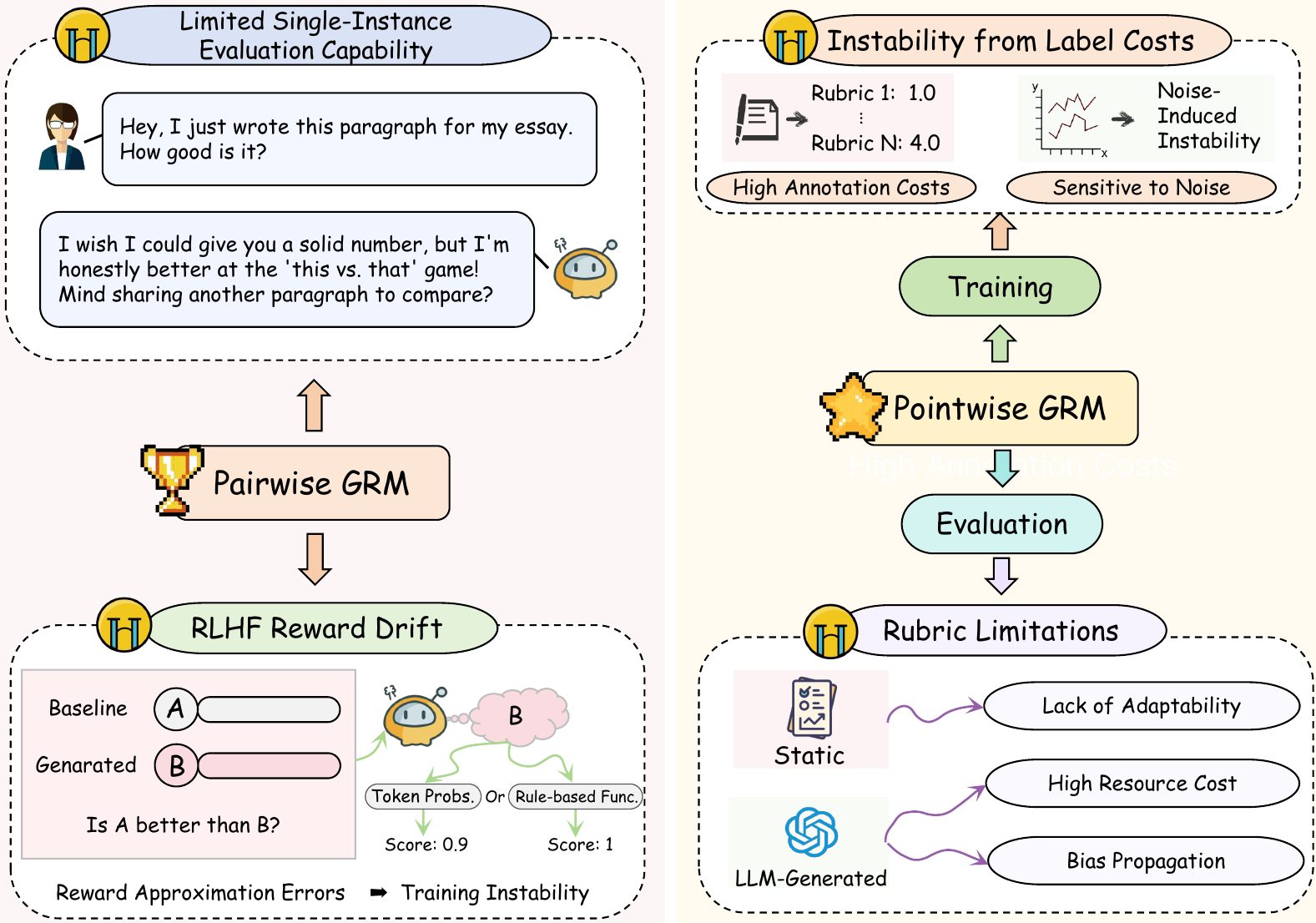}
	\caption{Challenges in two GRM Paradigms.}
	\label{fig:weak_grm}
	\vspace{-3ex}
\end{figure}

Current GRM training paradigms can be broadly categorized into two types.  
The first is \textbf{pairwise GRM}, which optimizes preference objectives by directly comparing response pairs. Although effective for relative preferences, it has two key limitations. 
First, it cannot handle single-instance evaluation tasks, as its inference
requires comparative inputs, limiting applicability to settings requiring
absolute quality assessment.
Second, the pairwise paradigm disrupts the RLHF pipeline~\citep{zhang2025reasoner,zhang2025continue} by converting comparative rewards into absolute ones, which introduces approximation errors and makes training less stable than
pointwise methods~\citep{xu2025unified}.

The second is \textbf{pointwise GRM}, which faces critical limitations in both evaluation and training phases. 
For evaluation, pointwise GRMs use static rubrics, utilizing either predefined rules that lack adaptability~\citep{promethueus,prometheus2} or LLM-generated criteria that incur high costs and bias risks~\citep{checklist,rar}.
In training, pointwise methods rely on costly, noise-prone absolute ratings for each rubric dimension, which, combined with unstable optimization dynamics, lead to high annotation costs and poor robustness.
As shown in Figure~\ref{fig:weak_grm}, these limitations highlight a core challenge in GRM design:
\textit{Can \textbf{pointwise GRMs} be trained with \textbf{adaptive rubrics} but \textbf{without explicit pointwise labels}?}

To address these challenges, we introduce the \textbf{P}reference-\textbf{a}ware \textbf{T}ask-\textbf{a}daptive \textbf{R}eward \textbf{M}odel (\ourtitle), a unified framework that enables pointwise GRM training directly from pairwise data without requiring explicit absolute labels. 
\ourtitle~integrates two core mechanisms. First, the \textbf{Preference-Aware Reward (PAR)} mechanism converts pairwise preferences into robust pointwise training signals, ensuring that selected responses consistently receive higher rubric-based scores than rejected ones. Second, \textbf{Dynamic Rubric Adaptation} generates context-aware, instance-specific evaluation criteria, overcoming the limitations of static rubrics and enabling precise alignment with diverse task requirements. Together, these mechanisms combine the data efficiency of pairwise training with the inference speed and interpretability of pointwise models. Rather than optimizing solely for static discriminative accuracy, \ourtitle~is designed to improve reward quality for generalization and downstream policy optimization.

In summary, our contributions are as follows:  
\begin{enumerate}[itemsep=1pt, parsep=0pt, topsep=0pt]
	\item We propose \textbf{\ourtitle}, a unified framework that integrates a \textbf{PAR mechanism} with \textbf{dynamic rubric adaptation}. PAR transforms pairwise preferences into robust pointwise training signals, enabling stable optimization without requiring explicit absolute labels.
	
	\item We introduce a \textbf{dynamic rubric adaptation mechanism} that generates both task-level and instance-specific evaluation criteria, overcoming the rigidity of static rubrics and enabling precise, context-aware assessment.
	
	\item Extensive experiments reveal a clear trade-off: while scalar BT models retain an advantage on static RewardBench pointwise accuracy, \textbf{\ourtitle} generalizes better on RMBench and provides stronger downstream utility under matched-data comparisons. Against identically trained BT-Qwen3 baselines, it consistently improves Best-of-N reranking, and when applied to downstream RLHF, it delivers an average relative improvement of \textbf{13.6\%} over the corresponding base policies on IFEval and InfoBench.
\end{enumerate}

\section{Related Work}

\textit{\textbf{Training Paradigms for Reward Modeling.}}
Reward modeling for RLHF primarily adopts either \textbf{pairwise} or \textbf{pointwise} supervision.
Pairwise training, such as the Bradley-Terry (BT) model~\citep{skywork,internlm2,eurus}, efficiently learns preferences from comparative judgments and supports single-instance evaluation in scalar models~\citep{sync}. However, many pairwise generative reward models require comparative inputs during both training and inference, limiting downstream flexibility~\citep{dspair,gram,rrm}.
Pointwise training relies on absolute scoring or rubric-based labeling for each response~\citep{promethueus,rar,QA}, enabling interpretable evaluations but incurring high annotation costs and demanding adaptive rubric design~\citep{Cloud,spct}. These limitations are especially pronounced in open-ended tasks with ambiguous evaluation criteria.

\textit{\textbf{Inference Paradigms for Reward Modeling.}}
The inference capabilities of reward models can be grouped into three main types.  
\textbf{Scalar RMs}, output numerical scores for single-instance evaluation, but often lack interpretability and fail to capture nuanced preferences in complex tasks~\citep{nexttoken}.
\textbf{Pointwise GRMs} provide rubric-based or reasoning-driven assessments for individual responses~\citep{promethueus,rar,rrm}, offering transparency but typically relying on costly explicit labels and static rubrics~\citep{spct,prometheus2}.
\textbf{Pairwise GRMs} focus on comparative assessment between response pairs~\citep{gram,genrm,rewardanything}, which restricts their use for absolute evaluation and complicates RLHF integration.

\textit{\textbf{Challenges in Bridging Training and Inference Gaps.}}
Recent work has sought to bridge these paradigms by combining pairwise and pointwise supervision~\citep{rewardanything,prometheus2,minigeneral} or using external models for rubric generation~\citep{rar}. However, these methods often incur additional computational costs and annotation burdens. The key challenge remains: efficiently training interpretable and adaptable pointwise generative reward models without costly explicit labels. Our approach addresses this by leveraging pairwise preference signals and dynamic rubric adaptation, effectively bridging the gap in RLHF reward modeling.

\begin{figure*}[t]
	\centering
	\includegraphics[width=0.88\textwidth]{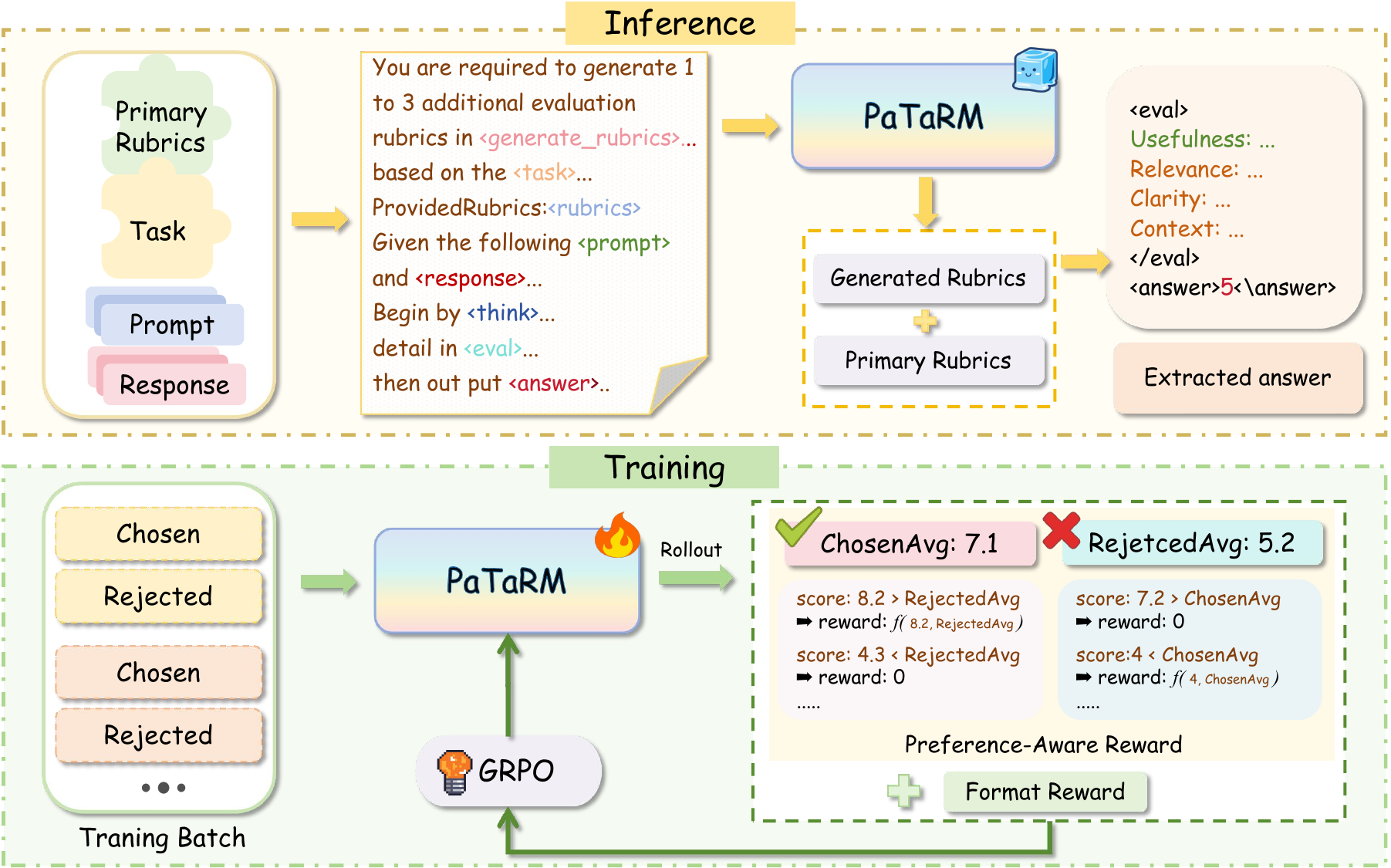}
	\caption{Overview of \ourtitle. The upper panel illustrates adaptive rubric generation, while the lower panel depicts the pointwise training procedure incorporating PAR and dynamic rubric adaptation.}
	\label{fig:framework}
\end{figure*}

\section{Methodology}
Figure~\ref{fig:framework} presents the overall pipeline of \ourtitle, which enables pointwise GRM training from pairwise data through two core mechanisms: \textbf{PAR} mechanism and \textbf{Dynamic Rubric Adaptation}. PAR transforms relative preference signals into robust pointwise training objectives, while dynamic rubrics generate context-aware evaluation criteria tailored to each instance.

\subsection{Preference-Aware Reward Mechanism}
\label{subsec:par}

Traditional reward modeling relies on either expensive absolute labels or pairwise comparisons that suffer from training-inference mismatch. 
We propose a preference-aware reward mechanism that enables pointwise training directly from pairwise data through generative evaluation.

\paragraph{Generative Judgment and Scoring.}
PaTaRM is designed as a generative reward model that, given a prompt $x$ and a pair of candidate responses (chosen $y^c$ and rejected $y^r$), produces $n$ judgment rollouts $\{y^c_i\}_{i=1}^n$ and $\{y^r_j\}_{j=1}^n$ based on the adaptive rubrics defined in Section~\ref{subsec:rubrics}. 
These rollouts yield individual scores $s^c_i$ and $s^r_j$, which are aggregated into averages $\bar{s}^c = \frac{1}{n} \sum_{i=1}^{n} s^c_i$ and $\bar{s}^r = \frac{1}{n} \sum_{j=1}^{n} s^r_j$ for the subsequent PAR calculation.

\paragraph{Optimization Objective.}
Our objective ensures that chosen responses consistently receive higher average scores than rejected ones, i.e., \( \bar{s}^c > \bar{s}^r \). This formulation enables end-to-end training with policy gradient methods (e.g., GRPO~\citep{r1,zhang2025rediscovering}, Reinforce++~\citep{reinforce++}, DAPO~\citep{dapo}) without the need for absolute golden scores.

\paragraph{Preference-Aware Reward Assignment.}  
For each rollout, the reward is assigned based on whether it satisfies the preference constraint:
\begin{equation*}
	\begin{aligned}
		R_{\text{PAR}}(y^c_i)
		&= \mathbb{I}\big[s^c_i > \bar{s}^r\big] \cdot f(\delta_i^c), \\
		R_{\text{PAR}}(y^r_j)
		&= \mathbb{I}\big[s^r_j <\bar{s}^c\big] \cdot f(\delta_j^r),
	\end{aligned}
\end{equation*}
where $\delta_i^c = s^c_i - \bar{s}^r$ and $\delta_j^r = \bar{s}^c - s^r_j$ denote the score margins, $\mathbb{I}[\cdot]$ is the indicator function, and $f(\cdot)$ maps the margin to a reward magnitude. 
This mechanism ensures that PaTaRM consistently ranks preferred responses higher than rejected ones, using only relative preference data. The formulation flexibly supports both binary and graded reward assignments, depending on the choice of $f(\cdot)$.

\paragraph{Format Reward.} 
To ensure well-formed outputs, we add a format penalty:
$$
R_{\mathrm{format}}(y) =
\begin{cases}
	-0.5,  & \text{if tags are incorrect}, \\
	-1.0, & \text{if score invalid}, \\
	0, & \text{otherwise.}
\end{cases}
$$
The total reward is $R(y|x) = R_{\mathrm{PAR}}(y|x) + R_{\mathrm{format}}(y)$.

\subsection{Dynamic Rubric Adaptation}
\label{subsec:rubrics}

Static rubrics limit adaptability and can lead to reward hacking. We introduce dynamic rubric adaptation that generates flexible, context-aware criteria by combining \textbf{global task-consistent criteria} with \textbf{instance-specific criteria} tailored to each prompt.

\paragraph{Rubric Generation.}
For each prompt $x$ and candidate response $y$, PaTaRM constructs the evaluation rubric $\mathcal{R}(x, y)$ by combining both global and instance-specific criteria. The global rubric provides a baseline for universal standards, while the instance-specific rubric adapts to the unique requirements and context of each example.

\paragraph{Rubric-Guided Scoring.}
During judgment rollouts, each response is evaluated according to its rubric $\mathcal{R}(x, y)$. The reward model produces a score $s(y)$ for response $y$ by aggregating its performance across all criteria. Unlike traditional approaches that require explicit manual assignment of criterion weights, PaTaRM leverages the inherent reasoning and balancing capabilities of LLMs to implicitly balance the importance of different criteria during evaluation. 
This enables more nuanced and context-aware scoring without the need for handcrafted weights, where previous work by \citep{rar} has validated the implicit weights can lead to better performance.

\subsection{Training Pipeline}

Our training consists of two stages:

\textbf{(1) Supervised Fine-Tuning (SFT):} We perform SFT on pointwise corpora constructed from pairwise data (see Appendix~\ref{sec:data_construction}), with further results detailed in Appendix~\ref{app:sft_analysis}.

\textbf{(2) Reinforcement Learning (RL):} We optimize the model using GRPO with group-relative advantages derived from PAR. This stabilizes learning by comparing responses within the same prompt group, eliminating the need for absolute labels.
Implementation details can be found in Appendix~\ref{sec:implementation}.

\begin{table*}[htbp]
	\centering
	\caption{
		Results on RewardBench and RMBench. 
		Models are grouped by family size to facilitate direct comparison between the Base model, Scalar RM (BT), and Generative RM (PaTaRM).
		$^{\dagger}$ denotes potential data contamination.
		$^{\ddagger}$ indicates reported performance.
	}
	\label{tab:results_combined}
	\setlength{\tabcolsep}{3pt}
	\renewcommand{\arraystretch}{1.15}
	
	\newcommand{\graycell}[1]{\textcolor{gray}{#1}}
	\newcommand{\best}[1]{\textbf{#1}}
	\newcommand{\second}[1]{\underline{#1}}
	
	\definecolor{bgref}{RGB}{255, 255, 220} 
	\definecolor{bg8b_title}{RGB}{255, 245, 245} 
	\definecolor{bg8b_body}{RGB}{255, 252, 252} 
	
	\definecolor{bg14b_title}{RGB}{255, 235, 235} 
	\definecolor{bg14b_body}{RGB}{255, 248, 248} 
	\resizebox{0.92\textwidth}{!}{
		\small
		\begin{tabular}{lcccccccccc}
			\toprule
			\multirow{2}{*}{\textbf{Model}} & \multicolumn{5}{c}{\textbf{RewardBench}} & \multicolumn{4}{c}{\textbf{RMBench}} \\
			\cmidrule(lr){2-6} \cmidrule(lr){7-10}
			& Overall & Chat & ChatHard & Safe & Reas. 
			& Overall & Easy & Medi. & Hard \\
			\midrule
			
			\rowcolor{gray!8}\multicolumn{10}{l}{\textit{General-purpose LLMs}} \\
			Gemini-1.5-Flash  & 73.1 & 90.7 & 60.8 & 78.7 & 62.3 & 51.3 & 66.4 & 50.3 & 37.4 \\
			DeepseekV3 & 75.2 & 85.8 & 59.0 & 75.2 & 80.9 & 51.2 & 66.9 & 50.0 & 36.8 \\
			GPT-4o  & 79.0 & 89.7 & 66.9 & 85.1 & 74.5 & 60.6 & 74.2 & 60.3 & 47.4 \\[0.5ex]
			
			\rowcolor{bgref}\multicolumn{10}{l}{\textit{Scalar Reward Models}} \\
			Skywork-Llama-3.1-8B$^{\dagger\ddagger}$  & {92.5} & {95.8} & 87.3 & 90.8 & {96.2} & 70.1 & {89.0} & 74.7 & 46.6 \\
			Skywork-Gemma-2-27B$^{\dagger\ddagger}$  & {93.8} & {95.8} &  {91.4} & {91.9} & {96.1} & 67.3 & 78.0 & 69.2 & 54.9 \\[0.5ex]
			
			\midrule
			
			\rowcolor{bg8b_title}
			\multicolumn{10}{l}{\textit{\textbf{Qwen3-8B Family}}} \\
			
			\rowcolor{bg8b_body}
			\hspace{0.5em}\graycell{Qwen3-8B \textit{(Base)}} & 
			\graycell{78.1} & \graycell{84.1} & \graycell{62.7} & \graycell{82.4} & \graycell{\second{83.2}} & 
			\graycell{\second{71.0}} & \graycell{79.5} & \graycell{\second{70.8}} & \graycell{\second{62.8}} \\
			
			\rowcolor{bg8b_body}
			\hspace{0.5em}BT-Qwen3-8B \textit{(Scalar)} &  
			\textbf{86.3} & \textbf{96.4} & \textbf{79.6} & \textbf{87.4} & 82.0 & 
			70.3 &\textbf{84.6} & 70.1 & 56.2  \\

			\rowcolor{bg8b_body}
			\hspace{0.5em}PaTaRM Qwen3-8B \textit{(Ours)}  & 
			\second{84.3} & \second{87.7} & \second{74.3} & \second{87.2} & \textbf{87.8} & 
			\textbf{78.7} & \second{82.8} & \textbf{78.7} & \textbf{74.7} \\[0.5ex]

			\rowcolor{bg14b_title}
			\multicolumn{10}{l}{\textit{\textbf{Qwen3-14B Family}}} \\
			
			\rowcolor{bg14b_body}
			\hspace{0.5em}\graycell{Qwen3-14B \textit{(Base)}} & 
			\graycell{81.9} & \graycell{87.4} & \graycell{69.3} & \graycell{84.6} & \graycell{86.2} & 
			\graycell{\second{73.2}} & \graycell{81.0} & \graycell{\second{73.8}} & \graycell{\second{64.9}} \\
			
			\rowcolor{bg14b_body}
			\hspace{0.5em}BT-Qwen3-14B \textit{(Scalar)} & 
			\textbf{89.9} & \textbf{95.3} & \textbf{87.5} & \second{87.6} & \second{89.2} & 
			70.9 & \second{85.8} & 70.7 & 56.2 \\

			\rowcolor{bg14b_body}
			\hspace{0.5em}PaTaRM Qwen3-14B \textit{(Ours)}  & 
			\underline{87.2} & \underline{91.5} & \second{77.9} & \textbf{87.8} & \textbf{91.5} & 
			\textbf{80.3} & \textbf{86.9} & \textbf{81.0} & \textbf{73.0} \\
			
			\bottomrule
		\end{tabular}
	}
\end{table*}

 \section{Experiment}
 \renewcommand{\thefootnote}{\fnsymbol{footnote}} 

 \subsection{Experiment Setup}
\paragraph{Reward Model Baselines.}
 We primarily adopt Qwen3~\citep{qwen3} as our base model. For comparison, we include three categories of baselines:
 
 \noindent\textbf{(1) Scalar RMs.} 
 These models replace the final projection layer with a scalar scoring head to output numerical preference scores. 
 We compare against the Skywork series~\citep{skywork} as we mainly use a subset of their training datasets.
To ensure a controlled comparison, we also train our BT-Qwen3 baselines using the identical dataset employed by PaTaRM.
 
 \noindent\textbf{(2) Pairwise GRMs.} 
 These models take a pair of responses as input to output a comparative judgment. 
 RRM~\citep{rrm} frames reward modeling as a reasoning task. RM-R1~\citep{rmr1} divides tasks into chat and reasoning types, where reasoning tasks require the model to first solve the problem. R3~\citep{r3} is an SFT-based series with integrated rubric generation.
 
 \noindent\textbf{(3) General-purpose LLMs.} 
 We also include proprietary systems such as GPT-4o~\citep{gpt4o}, Gemini-1.5-Flash and Gemini-1.5-Pro~\citep{gemini1.5pro}, and DeepseekV3~\citep{deepseekv3} as references.
 
 \paragraph{RLHF Baselines.}
In our downstream RLHF, we use Qwen2.5-7B, Qwen2.5-7B-Instruct, Qwen3-8B, and Qwen3-14B as policy models. All models are trained on the filtered dataset provided by RLCF~\citep{checklist}, which was constructed from Wildchat~\citep{wildchat}. 
For RL, we conduct GRPO using the PaTaRM-8B model as the reward model. As baselines, we include both SFT and DPO~\citep{dpo} trained on the same dataset, as well as GRPO guided by Skywork-Llama-3.1-8B. 
For brevity, we refer to the Skywork-Llama-3.1-8B model simply as Skywork throughout our downstream experiments.
 
 \paragraph{Evaluation.}
We evaluate RM and RLHF downstream task performance using their respective benchmark datasets. For RM, we use \textbf{RewardBench}~\citep{rewardbench}, which contains about 3,000 preference pairs across four domains, focusing on challenging cases requiring fine-grained alignment. 
In addition,  \textbf{RMBench}~\citep{rmbench} provides 1,300 preference pairs in \textit{chat}, \textit{math}, \textit{code}, and \textit{safety}, with stylistic variants and three difficulty levels (\textit{easy}, \textit{medium}, \textit{hard}), enabling robust evaluation.

For downstream RLHF tasks, we use \textbf{IFEval}~\citep{ifeval}, which evaluates instruction following on 541 prompts across 25 types of verifiable constraints, enabling systematic assessment. We also use \textbf{InfoBench}~\citep{infobench}, which contains 500 instructions and 2,250 decomposed evaluation questions across five categories, and uses ratiometric scoring over decomposed requirements for fine-grained, constraint-level analysis and efficient automated evaluation. For direct reward evaluation, we additionally perform Best-of-N reranking on AlpacaEval and MT-Bench, and we report reasoning-intensive RLHF results on Math-500 and GSM-8K.

\subsection{Results on Reward Model Benchmarks}

Table~\ref{tab:results_combined} presents the comparative evaluation on RewardBench and RMBench.
The prompt used for the untrained models is shown in Appendix~\ref{sec:prompt_general}.
Our analysis yields three key insights:

\paragraph{General-purpose LLMs $\neq$ Effective Reward Models.}
General-purpose models (e.g., GPT-4o), despite their strong instruction-following capabilities, lag significantly behind specialized models on discriminative benchmarks. This underscores that general pre-training is insufficient for fine-grained preference distinction, necessitating dedicated reward modeling.

\paragraph{Fragility and Data Hunger of Scalar Models.}
The {Scalar Reward Models} block reveals two critical limitations of the traditional BT paradigm. 

First, we observe a tendency for \textbf{distributional overfitting}. While Skywork achieves strong performance on RewardBench, it struggles significantly on RMBench, suggesting that it sacrifices general reasoning capabilities to fit the specific RewardBench distribution. 

Second, scalar models exhibit a severe \textbf{data scalability bottleneck}. 
Despite leveraging a stronger backbone architecture, the~{BT-Qwen3-8B} baseline achieves lower performance than Skywork on RewardBench.
We attribute this to data scale, as our models were trained on a curated subset rather than a massive corpus. This confirms that scalar models are highly data-hungry and require extensive data scaling to saturate performance.

\paragraph{PaTaRM vs. BT: A Trade-off under Controlled Data.}
Given the data dependency established above, the fairest comparison is between {PaTaRM} and reproduced {BT-Qwen3}, as they share the \textbf{identical training data distribution and volume}. 

Under this matched-data setting, BT-Qwen3 retains an advantage on static pointwise RewardBench, whereas PaTaRM generalizes markedly better to RMBench and subsequent downstream evaluations. In particular, the scalar BT models exhibit signs of \textbf{negative transfer}: both BT-Qwen3-8B and BT-Qwen3-14B underperform their respective unaligned Base models on RMBench Overall, indicating that optimizing for static discriminative accuracy can compromise broader reasoning ability. In contrast, relative to the corresponding base models, PaTaRM-Qwen3-8B achieves gains of \textbf{7.9\%} on RewardBench and \textbf{10.8\%} on RMBench, while the 14B model shows similar gains of \textbf{6.5\%} and \textbf{9.7\%}, respectively. This pattern suggests that PaTaRM better preserves the generalization and reward quality needed for downstream policy guidance.

\subsection{Best-of-N Re-ranking Evaluation}
To eliminate the model-family confounder in downstream evaluation, we conduct a Best-of-N (BoN) re-ranking experiment using identically trained BT-Qwen3 baselines. For each prompt, we sample 8 candidates (\texttt{temperature=1.0}, \texttt{max\_tokens=1024}) from either GPT-4o-mini or Qwen3-4B, and then use each reward model to select the highest-scoring response. We evaluate the selected responses on AlpacaEval (win rate against GPT-4.1-nano), MT-Bench, InfoBench, and IFEval; pairwise judgments are scored with GPT-4.1~\citep{gpt41}.

\begin{table}[t]
	\centering
	\caption{Best-of-N (N=8) re-ranking with identically trained BT-Qwen3 baselines. \textit{Average(Random)} denotes random selection among the 8 candidates.}
	\label{tab:bon_rerank}
	\setlength{\tabcolsep}{3.5pt}
	\renewcommand{\arraystretch}{1.12}
	\scriptsize
	\resizebox{\columnwidth}{!}{
	\begin{tabular}{lccccc}
		\toprule
		\textbf{Reward Model} & \textbf{AlpacaEval} & \textbf{MT-Bench} & \textbf{InfoBench} & \textbf{IFEval-P} & \textbf{IFEval-I} \\
		\midrule
		\rowcolor{gray!8}\multicolumn{6}{l}{\textit{Generator: GPT-4o-mini}} \\
		Average(Random) & 57.93 & 8.81 & 88.55 & 78.74 & 84.17 \\
		BT-Qwen3-8B & 59.88 & 9.07 & 89.42 & 80.41 & 85.73 \\
		PaTaRM-8B (Ours) & 59.94 & 9.13 & 89.86 & 79.85 & 85.85 \\
		BT-Qwen3-14B & 60.31 & 9.09 & 89.57 & 80.59 & 85.85 \\
		PaTaRM-14B (Ours) & \textbf{61.18} & \textbf{9.16} & \textbf{90.58} & \textbf{81.15} & \textbf{86.81} \\
		\midrule
		\rowcolor{gray!8}\multicolumn{6}{l}{\textit{Generator: Qwen3-4B}} \\
		Average(Random) & 14.21 & 7.04 & 84.06 & 66.54 & 74.82 \\
		BT-Qwen3-8B & 14.66 & 7.06 & 84.20 & 76.52 & 82.49 \\
		PaTaRM-8B (Ours) & 19.99 & 7.58 & 87.05 & \textbf{78.56} & 83.93 \\
		BT-Qwen3-14B & 15.65 & \textbf{7.72} & 85.80 & 75.42 & 81.89 \\
		PaTaRM-14B (Ours) & \textbf{20.30} & 7.59 & \textbf{87.83} & \textbf{78.56} & \textbf{84.17} \\
		\bottomrule
	\end{tabular}
	}
\end{table}

Results in Table~\ref{tab:bon_rerank} directly validate reward quality under matched-data comparisons. On the weaker Qwen3-4B generator, PaTaRM-8B improves AlpacaEval win rate by \textbf{5.33\%} over BT-Qwen3-8B and improves InfoBench by \textbf{2.85\%}, showing that it more reliably selects high-quality outputs from noisy candidate sets. Even on the stronger GPT-4o-mini generator, PaTaRM-14B outperforms BT-Qwen3-14B across all metrics. These results indicate that PaTaRM's advantage stems from the reward signal itself rather than from backbone mismatch.

 \subsection{RLHF Downstream Performance}
 Beyond re-ranking, we evaluate whether PaTaRM can guide actual policy optimization on two downstream settings: instruction following and reasoning-intensive tasks.
 
 \paragraph{Instruction-following tasks.}
 This task family is excluded from RM training, so the reward model must generalize in a zero-shot manner from the provided rubrics rather than from direct task-specific supervision (see Figure~\ref{fig:instruct}). We train policies with GRPO using PaTaRM and compare against SFT, DPO, and GRPO guided by Skywork.
 \footnotetext[2]{All GPT-4o results reported in our experiments are based on the {2024-08-06} version.}
   \begin{table}[h]
 	\centering
 	\caption{Main Comparative Analysis of Downstream RLHF Performance.}
 	\label{tab:main_downstream_task}
 	\setlength{\tabcolsep}{3pt}
 	\renewcommand{\arraystretch}{1.2} 
 	
 	\definecolor{bg7b}{RGB}{248, 252, 255}  
 	\definecolor{bg14b}{RGB}{240, 248, 255} 
 	
 	\resizebox{\columnwidth}{!}{
 		\begin{tabular}{lcccccccc}
 			\toprule
 			& \multicolumn{2}{c}{IFEval (Prompt)} & \multicolumn{2}{c}{IFEval (Inst.)} & \multirow{2}{*}[-2.6ex]{\textbf{Avg}}  & \multicolumn{3}{c}{InfoBench} \\
 			\cmidrule(lr){2-3} \cmidrule(lr){4-5} \cmidrule(lr){7-9}
 			\textbf{Model} & \textbf{Loose} & \textbf{Strict} & \textbf{Loose} & \textbf{Strict} & & \textbf{Easy} & \textbf{Hard} & \textbf{Overall} \\
 			\midrule
 			
 			GPT-4o  & 79.5 & 77.1 & 83.7 & 85.5 & 81.4 & 87.9 & 87.6 & 87.1 \\
 			\midrule
 			
 			\rowcolor{bg7b} Qwen2.5-7B-Base & 41.7 & 32.0 & 47.7 & 38.8 & 40.1 & 67.6 & 65.2 & 66.7 \\
 			\rowcolor{bg7b}
 			\quad + SFT              & 41.0 & 32.5 & 54.7 & 45.2 & 43.4 & 80.9 & 67.8 & 71.8 \\
 			\rowcolor{bg7b}
 			\quad + DPO (RLCF)            & 44.9 & 36.6 & 55.5 & \underline{48.1} & 46.3 & \textbf{85.6} & \underline{77.2} & \underline{79.8} \\
 			\rowcolor{bg7b}
 			\quad + RL w/ Skywork     & \underline{46.0} & \underline{36.8} & \underline{56.4} & 47.5 & \underline{46.7} & 77.1 & 73.6 & 78.7 \\
 			\rowcolor{bg7b}
 			\quad + RL w/ PaTaRM     & \textbf{48.1} & \textbf{38.1} & \textbf{60.2} & \textbf{50.4} & \textbf{49.2} & \underline{83.7} & \textbf{84.6} & \textbf{84.3} \\
 			
 			\midrule
 			
 			\rowcolor{bg14b} Qwen3-14B        & 88.2 & 85.8 & 91.8 & 90.3 & 89.0 & 86.3 & 86.9 & 86.7 \\
 			\rowcolor{bg14b} 
 			\quad + SFT              & 85.6 & 83.5 & 90.3 & 89.0 & 87.1 & 87.4 & 86.0 & 86.4 \\
 			\rowcolor{bg14b} 
 			\quad + DPO (RLCF)             & 88.7 & 85.8 & 92.6 & 90.6 & 89.4 & \underline{88.7} & 86.5 & 87.2 \\
 			\rowcolor{bg14b}  
 			\quad + RL w/ Skywork    & \underline{89.1} & \underline{86.5} & \underline{92.7} & \underline{91.0} & \underline{89.8} & 87.1 & \underline{88.1} & \underline{87.8} \\
 			\rowcolor{bg14b}  
 			\quad + RL w/ PaTaRM     & \textbf{90.2} & \textbf{87.8} & \textbf{93.7} & \textbf{92.1} & \textbf{90.9} & \textbf{89.2} & \textbf{89.2} & \textbf{89.2} \\
 			
 			\bottomrule
 		\end{tabular}
 	}
 	\vspace{-1mm}
 \end{table}
 As shown in Table~\ref{tab:main_downstream_task}, PaTaRM consistently drives the highest policy performance across both model scales. Relative to the corresponding base policies, PaTaRM yields substantial improvements on Qwen2.5-7B-Base, boosting IFEval scores by \textbf{22.7\%} and InfoBench scores by \textbf{26.4\%}. Even on the stronger Qwen3-14B, it achieves further gains of \textbf{2.1\%} and \textbf{2.9\%}, respectively.
 Compared with the corresponding SFT and DPO baselines, RL with PaTaRM yields larger and more stable gains, indicating that explicit generative reward modeling provides denser supervision than direct preference optimization alone. PaTaRM also outperforms Skywork in this zero-shot transfer setting, complementing the matched-data BoN results above and reinforcing our central claim: PaTaRM's main advantage lies in reward quality and downstream generalization rather than in maximizing a single static benchmark.

 \paragraph{Reasoning-intensive tasks.}
We further evaluate whether PaTaRM can guide policy learning on objective reasoning tasks rather than only open-ended chat or instruction-following tasks. Following the same training setup, we optimize policies on a merged GSM-8K/Math-500 training corpus (11,973 samples) and report performance at step 96 due to computational cost. As shown in Table~\ref{tab:math_reasoning_main}, PaTaRM consistently outperforms both rule-based rewards and Skywork-Llama-3.1-8B across both policy scales. The gains are especially pronounced for the weaker Qwen3-0.6B policy, suggesting that PaTaRM is particularly effective when the reward model must distinguish partially correct reasoning traces.

\begin{table}[htbp]
	\centering
	\caption{RLHF on reasoning-intensive tasks.}
	\label{tab:math_reasoning_main}
	\setlength{\tabcolsep}{4pt}
	\renewcommand{\arraystretch}{1.12}
	\resizebox{0.48\textwidth}{!}{
		\begin{tabular}{lcc}
			\toprule
			\textbf{Method} & \textbf{Math-500} & \textbf{GSM-8K} \\
			\midrule
			\rowcolor{gray!8}\multicolumn{3}{l}{\textit{Policy: Qwen3-8B}} \\
			Base Model & 90.0 & 89.8 \\
			+ Skywork-Llama-3.1-8B & 93.6 & 93.7 \\
			+ Rule-based Reward & 95.0 & 90.6 \\
			\textbf{+ PaTaRM (Ours)} & \textbf{95.2} & \textbf{94.3} \\
			\midrule
			\rowcolor{gray!8}\multicolumn{3}{l}{\textit{Policy: Qwen3-0.6B}} \\
			Base Model & 72.2 & 77.0 \\
			+ Skywork-Llama-3.1-8B & 74.2 & 77.3 \\
			+ Rule-based Reward & 76.4 & 78.4 \\
			\textbf{+ PaTaRM (Ours)} & \textbf{78.0} & \textbf{81.0} \\
			\bottomrule
		\end{tabular}
	}
\end{table}

 \subsection{Evaluation in Pairwise Setting}
 To further assess the robustness of PaTaRM, we evaluate it in a pairwise setting by applying the model directly to a pairwise inference template \textbf{without additional training}. 
 
 As presented in Table~\ref{tab:results_pairwise_purple}, despite only being trained via a pointwise paradigm, PaTaRM demonstrates remarkable adaptability. 
 At the 8B scale, it remains highly competitive with specialized pairwise models. More importantly, at the 14B scale, \textbf{PaTaRM outperforms all baselines, achieving the highest Overall score of 89.7}. 
 Crucially, PaTaRM consistently excels on the \textit{ChatHard} and \textit{Safety} subsets across both scales. 
 This suggests that our dynamic rubric mechanism captures granular preference distinctions and safety constraints more effectively than standard pairwise training, which tends to rely on holistic but vague impressions. This result confirms that PaTaRM learns a generalized and robust understanding of preference that transcends specific scoring formats.
 \begin{table}[htbp]
 	\centering
 	\caption{
 		Pairwise Inference on RewardBench. 
 	}
 	\label{tab:results_pairwise_purple}
 	\setlength{\tabcolsep}{3.5pt}
 	\renewcommand{\arraystretch}{1.10}
 	\small
 	
 	\definecolor{bg_purple_light}{RGB}{252, 248, 255} 
 	\definecolor{bg_purple_deep}{RGB}{242, 235, 255} 
 	
 	\resizebox{0.48\textwidth}{!}{
 		\begin{tabular}{lccccc}
 			\toprule
 			\textbf{Model} & \textbf{Overall} & \textbf{Chat} & \textbf{Chat.H} & \textbf{Safe} & \textbf{Reas.} \\
 			\midrule
 			
 			\rowcolor{gray!8}
 			\multicolumn{6}{l}{\textit{General-purpose LLMs}} \\
 			GPT-4o $^{\ddagger}$ & 86.7 & 96.1 & 76.1 & 86.6 & 88.1  \\
 			Gemini-1.5-Pro$^{\ddagger}$ & 88.2 & 92.3 & 80.6 & 87.9 & 92.0  \\
 			\midrule
 			
 			\rowcolor{bg_purple_light} \multicolumn{6}{l}{\textit{7B/8B Models}} \\
 			\rowcolor{bg_purple_light}
 			RRM-7B$^{\ddagger}$            & 82.2 & 87.7 & 70.4 & 80.7 & 90.0 \\
 			\rowcolor{bg_purple_light}
 			RM-R1 Qwen-7B$^{\ddagger}$     & 85.2 & \textbf{94.1} & 74.6 & 85.2 & 86.7 \\
 			\rowcolor{bg_purple_light}
 			R3-Qwen3-8B-14k$^{\ddagger}$   & \textbf{87.5} & \underline{93.3} & \underline{75.7} & \underline{85.7} & \textbf{95.3} \\
 			\rowcolor{bg_purple_light}
 			PaTaRM Qwen3-8B                & \underline{87.0} & 89.6 & \textbf{77.1} & \textbf{86.4} & \underline{95.1} \\
 			
 			\midrule
 			
 			\rowcolor{bg_purple_deep} \multicolumn{6}{l}{\textit{14B Models}} \\
 			\rowcolor{bg_purple_deep}
 			RM-R1 Qwen-14B$^{\ddagger}$    & \underline{88.2} & \textbf{93.6} & \underline{80.5} & \underline{86.9} & 92.0 \\
 			\rowcolor{bg_purple_deep}
 			R3-Qwen3-14B-14k$^{\ddagger}$  & \underline{88.2} & \textbf{93.6} & 77.6 & 85.3 & \textbf{96.3} \\
 			\rowcolor{bg_purple_deep}
 			PaTaRM Qwen3-14B               & \textbf{89.7} & \underline{93.2} & \textbf{82.6} & \textbf{87.5} & \underline{95.6} \\
 			\bottomrule
 		\end{tabular}
 	}
 \end{table}

\section{Analysis}

\subsection{Robustness to Noisy Labels}
\label{sec:noisy_robustness}

\begin{figure}[t]
	\centering
	
	\begin{subfigure}[b]{0.9\linewidth}
		\centering
		\includegraphics[width=\linewidth]{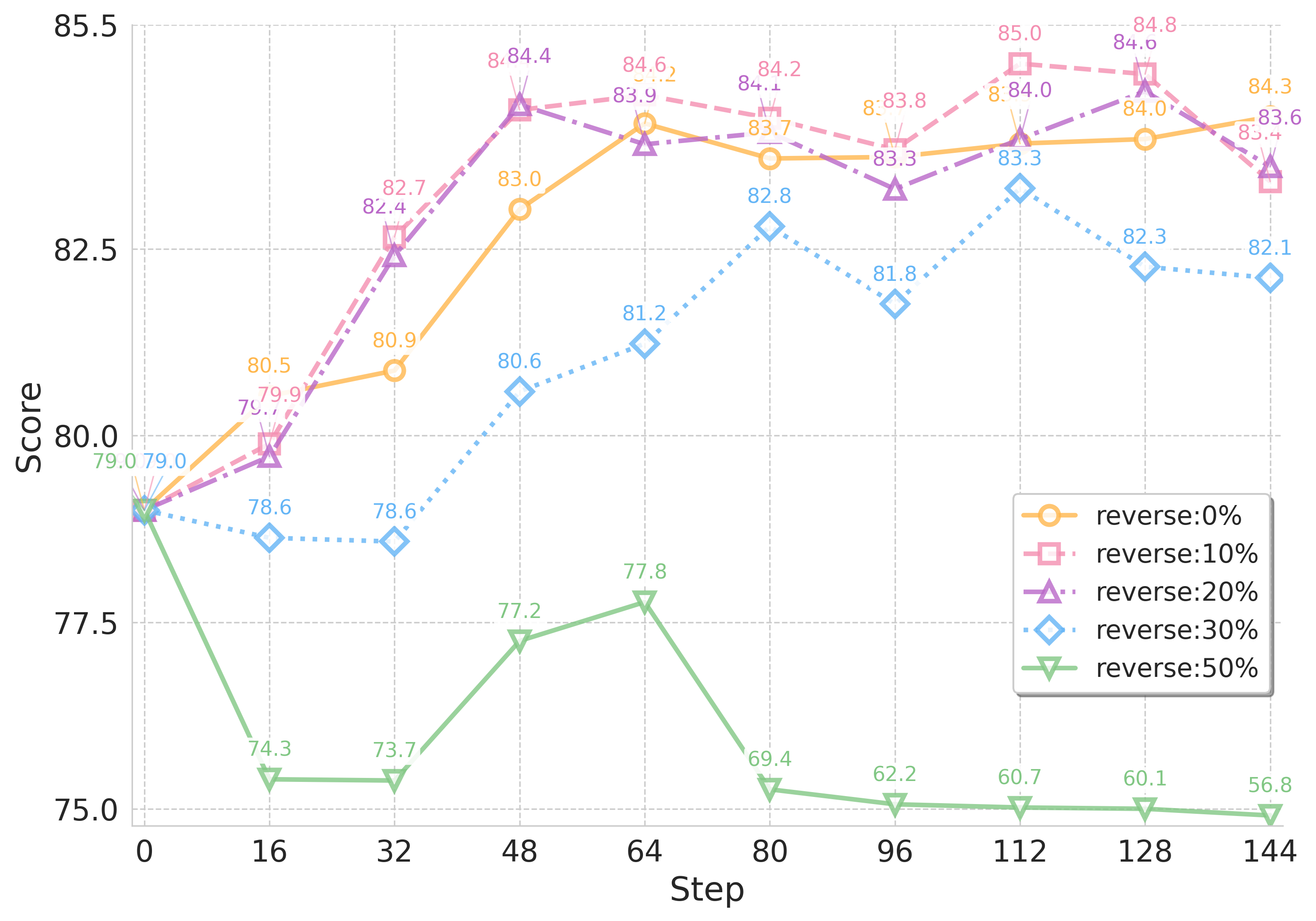} 
		\caption{The score of PaTaRM.}
	\end{subfigure}
	\hfill 
	\begin{subfigure}[b]{0.9\linewidth}
		\centering
		\includegraphics[width=\linewidth]{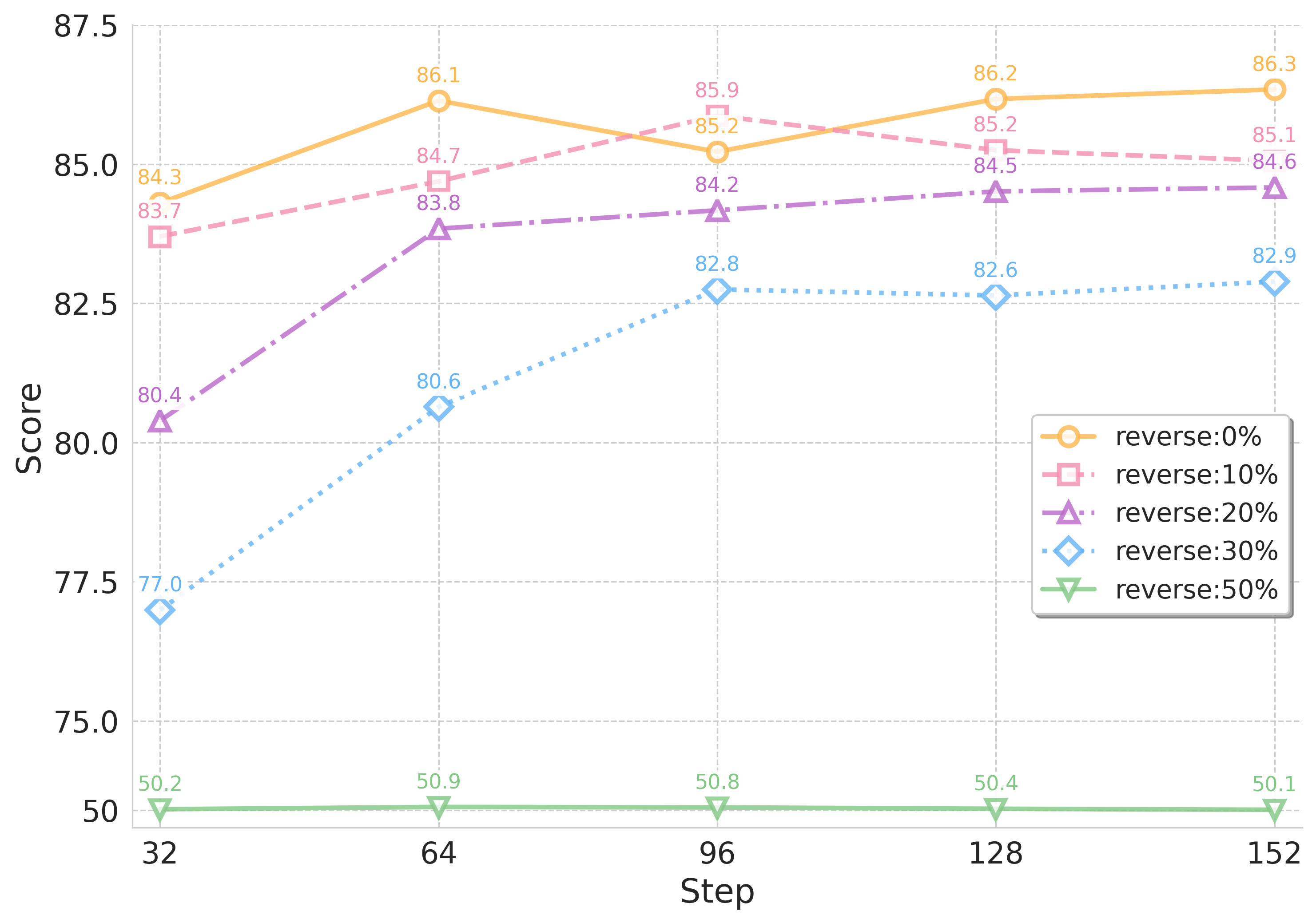}
		\caption{The score of BT-RM.}
	\end{subfigure}
	

	\caption{
		Noise-robustness comparison between PaTaRM and BT-RM on RewardBench.
	}
	\label{fig:reverse}
	\vspace{-1ex}
\end{figure}

To evaluate resilience against data corruption, we retrain BT-RM and PaTaRM on datasets with randomly flipped preference labels. Figure~\ref{fig:reverse} reveals distinct behaviors across noise regimes, highlighting the unique robustness of our approach.
This setting is realistic because large-scale preference corpora inevitably contain annotation inconsistencies and ambiguous comparisons.

\paragraph{Mitigating Shortcut Learning via Mild Noise (10-20\%).}
Surprisingly, in low-to-moderate noise regimes~(10-20\%), PaTaRM achieves a \textbf{higher peak performance} than the noise-free baseline. 
\textbf{In contrast, while BT fits the noisy distribution, its peak performance steadily declines.}

We attribute this counter-intuitive phenomenon to the mitigation of {shortcut learning}. 
In the absence of noise, the model may prematurely converge by exploiting superficial patterns or heuristics.
The introduction of mild noise (10-20\%) disrupts these brittle correlations, forcing the model to rely on deeper, more robust reasoning paths to satisfy the reward mechanism.
\paragraph{Dynamics of Resistance (50\% Noise).}
At the extreme noise level of 50\%, the difference between the two methods becomes stark. The BT model collapses to random performance because its scalar objective must fit mutually contradictory labels.
In contrast, PaTaRM exhibits a \textbf{distinct recovery trajectory} after the initial drop. 
We attribute this to PAR’s stochastic score aggregation and dense rollout-level credit assignment, where the model optimizes the relative dominance of chosen and rejected score distributions across multiple rollouts rather than fitting a single binary signal.
This acts as an implicit multi-sample voting mechanism, reduces gradient variance, and allows the model to preserve useful ranking structure inherited from the SFT even under severe corruption. 
Appendix~\ref{app:pairwise_noise} further shows that a standard pairwise GRM still collapses at 50\% noise, while Appendix~\ref{app:score_dynamics} confirms that PaTaRM does not rely on unbounded upward score drift to enlarge the margin.

Overall, scalar models are highly vulnerable to corrupted supervision, whereas PaTaRM remains substantially more robust because PAR converts sparse pairwise preferences into a lower-variance, distributional training signal.
\subsection{Ablation Study on Rubric Components}

To assess the contribution of different rubric components, we conducted an ablation study comparing our task-adaptive strategy with predefined rubrics and model-generated constraints. Given the instability observed in the baselines, we report the peak performance achieved during training in Table~\ref{tab:ablation}.

The \textbf{Only Primary} setting relies solely on static rules. We observed that this method reaches its peak performance early in the training steps, indicating a tendency to overfit to surface-level features and converge prematurely.
Conversely, the \textbf{Only Generated} setting relies exclusively on dynamic constraints. Without the grounding of predefined rules, this setting exhibits a distinct performance decline as training progresses. 
However, its superior peak performance on the {ChatHard} subset confirms that dynamic constraints are essential for capturing subtle preference distinctions that static rules miss.
Our \textbf{task-adaptive} approach achieves the best overall performance by synergizing these components. It uses primary rubrics as a stabilizing anchor, while leveraging generated constraints to introduce necessary variance, effectively balancing stability with adaptability.

\begin{table}[t]
	\centering
	\caption{
		Ablation study on rubric composition. 
		\textbf{Primary}: predefined rubrics; 
		\textbf{Generated}: model-generated constraints.
			\textit{task-adaptive} (Ours) achieves the best overall balance.
	}
	\label{tab:ablation}
	
	\definecolor{rowgray}{gray}{0.95}
	
	\resizebox{\columnwidth}{!}{
		\begin{tabular}{lccccc}
			\toprule
			\textbf{Rubric Setting} & \textbf{Avg} & \textbf{Chat} & \textbf{ChatHard} & \textbf{Safe} & \textbf{Reas.} \\
			\midrule
			
			Only Primary & 80.4 & \textbf{92.5} & 64.7 & 77.0 & \textbf{87.2} \\
			
			Only Generated & 81.2 & 88.6 & \textbf{70.4} & \textbf{82.0} & 83.8 \\
			
			\rowcolor{rowgray}
			\textbf{task-adaptive (Ours)} & \textbf{81.4} & 90.8 & 67.5 & 80.3 & 87.0 \\
			
			\bottomrule
		\end{tabular}
	}
\end{table}

\subsection[Does the Design of f(.) Matter?]{Does the Design of $f(\cdot)$ Matter?}
\label{sec:ablation_mapping}

As defined in Section~\ref{subsec:par}, $f(\cdot)$ determines how rewards are assigned based on the score margin between chosen and rejected responses. We investigate two instantiations of $f(\cdot)$.

\noindent
\textbf{Graded function ($f(\delta) = \Delta$).}  
We define $\Delta$ as a graded reward assignment: $f(\delta) = 1.2$ if $0 < \delta \leq 2$, and $f(\delta) = 1.4$ if $\delta > 2$. 
Here, $\delta$ denotes the score margin between chosen and rejected responses. 
This setting aligns with our SFT data filtering strategy, where a margin of $2$ serves as the threshold for reliable preference quality. 

\noindent
\textbf{Constant function ($f(\delta) = \alpha$).}  
We define $\alpha$ as a constant reward: $f(\delta) = 1.3$ for all $\delta > 0$, where any positive margin directly yields a fixed reward. This formulation simplifies the assignment and disregards the magnitude of preference gaps.

\begin{figure}[t]
	\centering
	
	\begin{subfigure}[b]{0.88\linewidth}
		\centering
		\includegraphics[width=\linewidth]{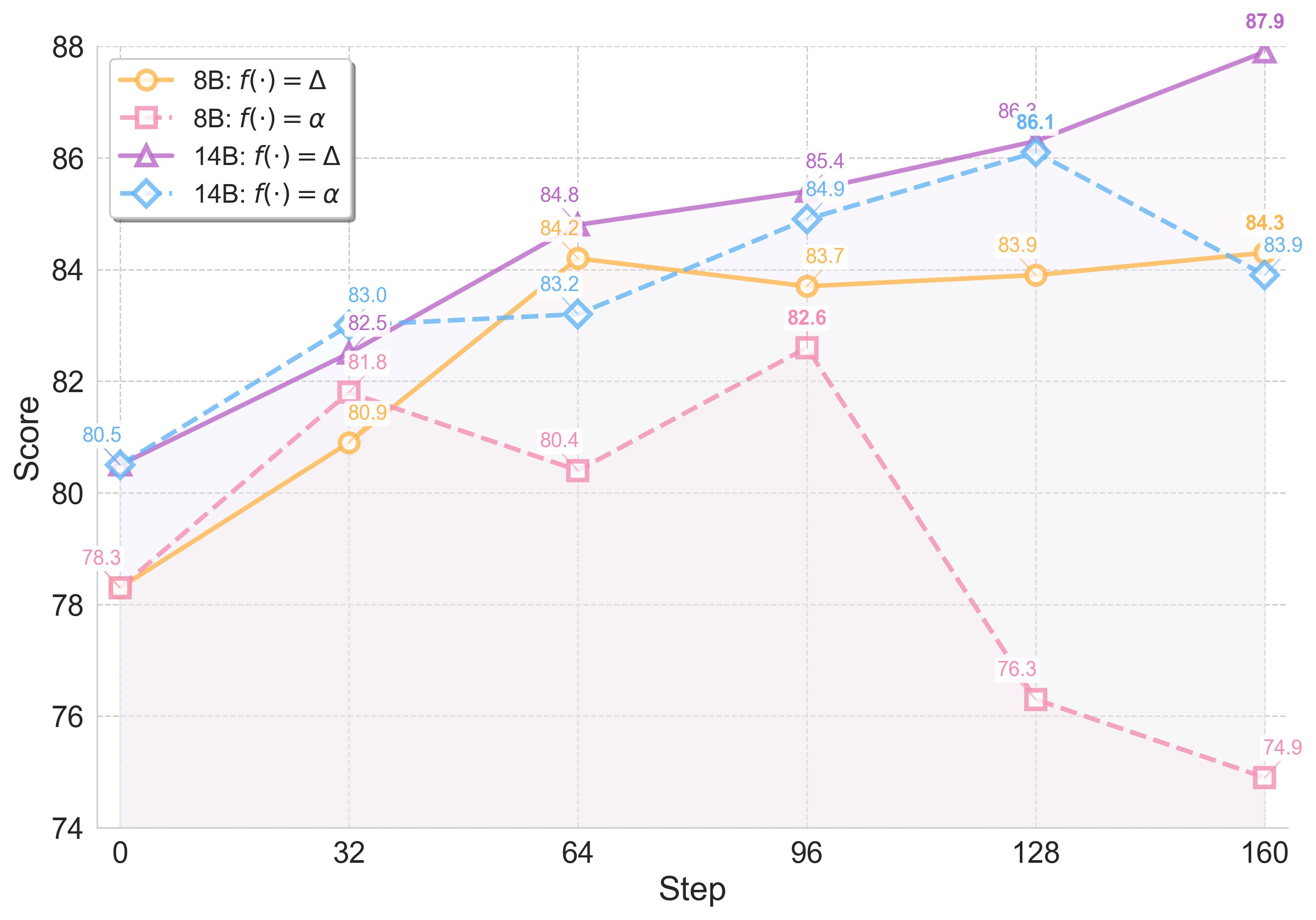} 
		\caption{RewardBench Results.}
	\end{subfigure}
	\hfill 
	\begin{subfigure}[b]{0.88\linewidth}
		\centering
		\includegraphics[width=\linewidth]{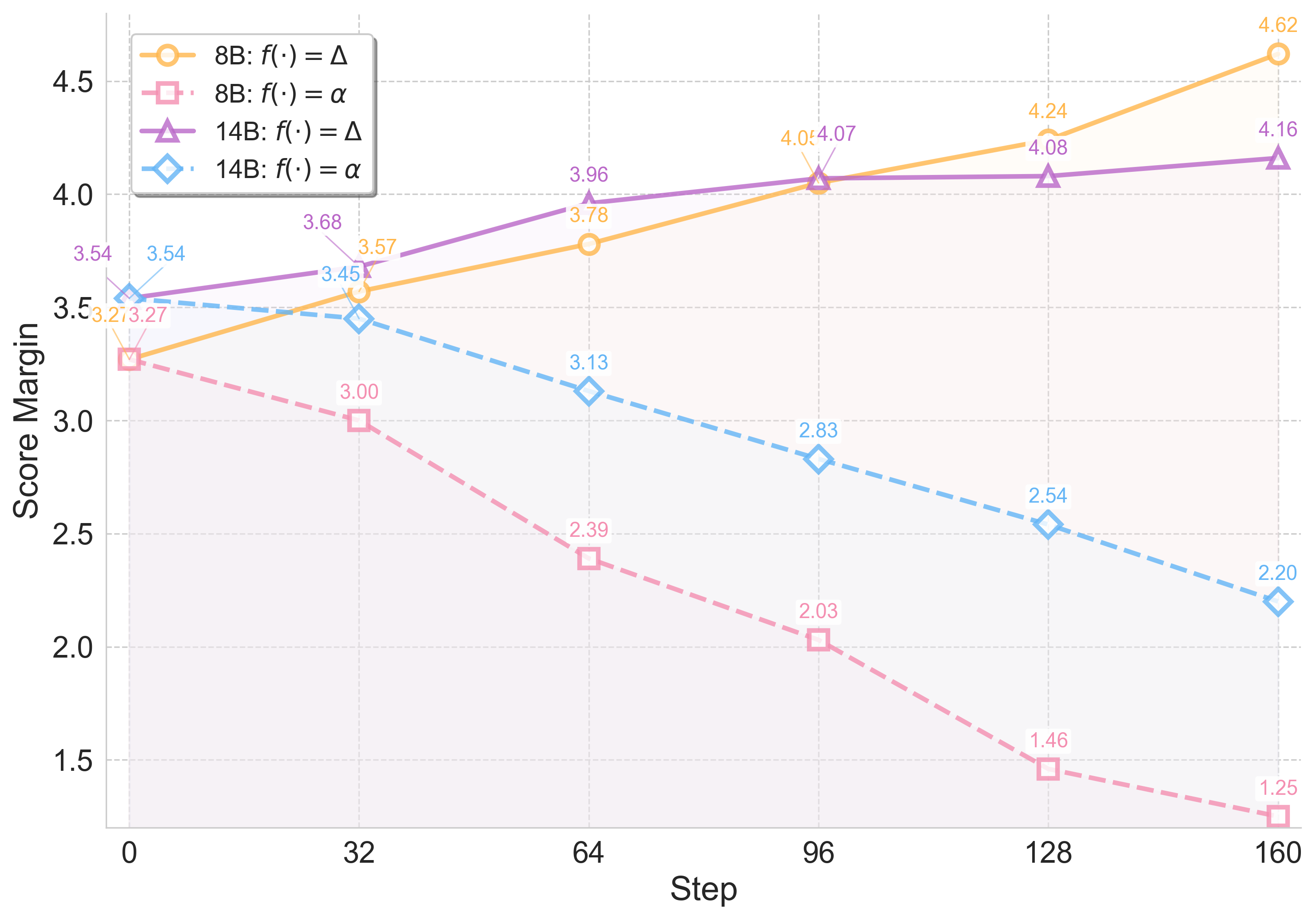}
		\caption{Average Score Margin.}
	\end{subfigure}
	
	\vspace{-1ex}
	
	\caption{
		Impact of reward functions $f(\cdot)$ across steps.
	}
	\label{fig:score_analysis}
	
	\vspace{-3ex}
\end{figure}

\medskip
\noindent
\textbf{Results and Analysis.}
Figure~\ref{fig:score_analysis} reveals a critical interaction between reward assignment and training stability. The 8B model trained with the constant $\alpha$ suffers a catastrophic performance collapse in later stages. This instability stems from \textbf{reward hacking} driven by the uncalibrated constant signal. Since the model receives the full reward $\alpha$ for even a marginal superiority, it is incentivized to exploit shortcuts rather than learning robust features that justify a larger semantic gap. 
This leads to \textbf{margin decay}, where the discriminative boundary becomes fragile and susceptible to noise. In contrast, the graded $\Delta$ provides a \textbf{dense reward signal} that aligns the reward magnitude with the rubric-defined quality gap, effectively regularizing the training and preventing such hacking behavior.

\subsection{Time Scaling Analysis}
For \textbf{scalar models}, voting is usually done by averaging the predicted scores of multiple outputs. However, because scalar values tend to have limited variance, this approach often struggles to scale and fails to capture subtle differences between responses~\citep{spct,Cloud}.  

\begin{figure}[htbp]
	\centering
	\includegraphics[width=0.9\linewidth]{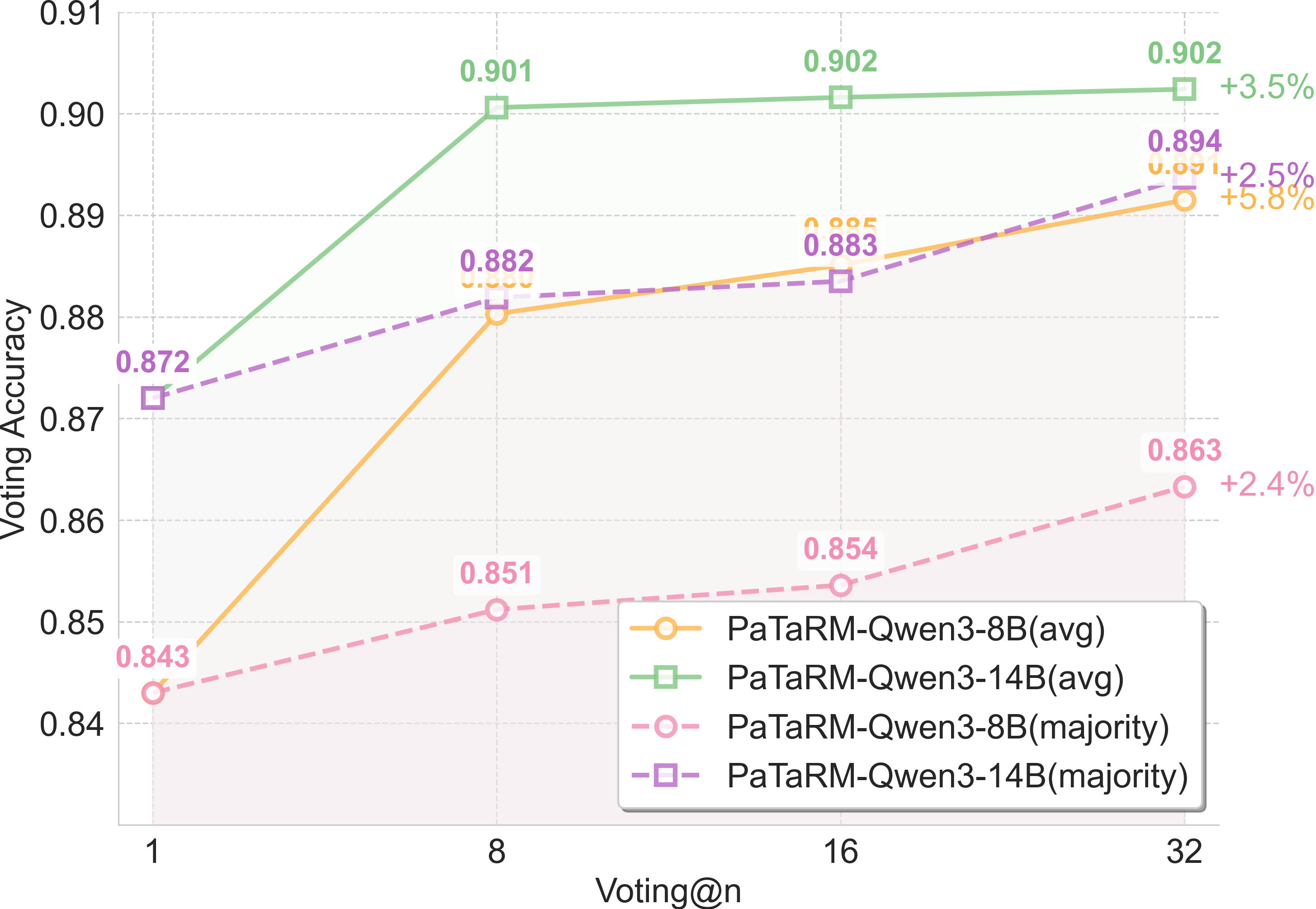}
	\caption{Performance of voting@n on RewardBench.}
	\label{fig:vote}
\end{figure}

For \textbf{pairwise GRMs}, voting adopts a majority rule, where the response most frequently preferred is selected as the best. This scales better with more samples but may introduce bias since ties are excluded and fine-grained distinctions are ignored~\citep{selftaught}.  
As shown in Fig~\ref{fig:vote}, we investigate PaTaRM under both voting schemes.  With \textbf{average voting}, the gains are particularly notable, showing clear benefits even at $n=8$, likely due to the PAR mechanism which strengthens mean-level improvements. With \textbf{majority voting}, the improvements are steadier but less sharp, reflecting a smoother scaling behavior. Overall, PaTaRM demonstrates robust advantages regardless of the voting strategy.

\section{Conclusions}

We introduce PaTaRM, a unified RLHF framework that enables training point-wise GRMs with adaptive rubrics without explicit point-wise labels, combining a preference-aware reward mechanism with dynamic rubric adaptation for efficient and interpretable pointwise reward modeling.
PaTaRM achieves the \textbf{$\mathcal{O}(N)$} efficiency of pointwise scoring while retaining the \textbf{data efficiency} of pairwise supervision, and maintains strong performance when transferring from \textbf{pairwise inference format}.
 
Extensive experiments on RewardBench and RMBench show that PaTaRM achieves an average relative improvement of \textbf{8.7\%} across the Qwen3-8B and Qwen3-14B models. Crucially, PaTaRM enhances downstream RLHF performance in out-of-domain settings, yielding substantial relative improvements up to \textbf{26.4\%} on Qwen2.5-7B-Base and \textbf{2.9\%} on Qwen3-14B across IFEval and InfoBench, respectively. 
Overall, PaTaRM establishes a solid foundation for advancing the development of more capable, generalizable, and interpretable reward models in reinforcement learning from human feedback.

\newpage

\section*{Limitations}
Our proposed method, PaTaRM, demonstrates significant improvements in reward modeling. 
However, several limitations remain. First, although we reduce the reliance on expensive scalar annotations, the quality of the pairwise preference data still fundamentally bounds performance. 
Second, while the generated reasoning provides transparency, we have not explicitly optimized for the faithfulness of these explanations to the model's internal decision-making process. 
Future work will focus on addressing these constraints.

\section*{Ethical Considerations}

	\textbf{Informed Consent:} All data collection processes involving human participants (if any) have obtained necessary informed consent.
	
	\textbf{Privacy Protection:} The datasets utilized in this study are derived from open-source repositories and adhere to privacy protection principles. We have verified that no personally identifiable information is exposed.
	
	\textbf{Bias Mitigation:} We have considered potential biases during the model design and evaluation phases. While reward models can inadvertently reinforce societal biases present in training data, we aim to mitigate this through diverse data sourcing.
	
	\textbf{Transparency:} Research funding sources are transparent, and there are no conflicts of interest to declare.

\section*{Reproducibility Statement}
To ensure the reproducibility of our results, we provide the following resources and details. All experiments reported in this paper can be reproduced using NVIDIA A100 GPUs:
\begin{enumerate}
	\item \textbf{Code, Data, and Checkpoints:} The complete resources are available at \url{https://huggingface.co/AIJian/PaTaRM}.
	\item \textbf{Data Processing:} The detailed dataset preprocessing pipeline is described in Appendix~\ref{sec:data_construction}.
	\item \textbf{Hyperparameters:} All model training hyperparameter configurations are listed in Table~\ref{tab:hyperparameters}.
	\item \textbf{Environment:} Hardware specifications and environmental setups are detailed in Appendix~\ref{sec:Training_setting}.
\end{enumerate}

\bibliography{custom}
\appendix
\clearpage

\section{Clarification of Different Reward Model Architectures}
\label{app:complexity_analysis}
\begin{table*}[b]
	\centering
	\caption{
		Comparison of Reward Model Architectures. 
		\textbf{Ranking Complexity} denotes the number of model forward passes required to rank $N$ candidate responses.
		PaTaRM uniquely combines the data efficiency of pairwise training with the inference efficiency of pointwise models.
	}
	\label{tab:model_comparison}
	\setlength{\tabcolsep}{5pt}
	\renewcommand{\arraystretch}{1.3}
	\resizebox{\textwidth}{!}{
		\begin{tabular}{lcccc}
			\toprule
			\textbf{Model Type} & \textbf{Training Data Format} & \textbf{Inference Input} & \textbf{Inference Output} & \textbf{Ranking Complexity} \\
			\midrule
			\textbf{BT Scalar Model} & 
			Pairs $(y_w, y_l)$ & 
			Single Response $y$ & 
			Scalar Score $s \in \mathbb{R}$ & 
			$\mathcal{O}(N)$ \\
			
			\textbf{Pairwise GRM} & 
			Pairs $(y_w, y_l)$ & 
			Response Pair $(y_A, y_B)$ & 
			Judgment (e.g., ``A $>$ B'') & 
			$\mathcal{O}(N \log N)$ \\
			
			\textbf{Pointwise GRM} & 
			Single $y$ + Rating/Score & 
			Single Response $y$ & 
			Evaluation Text / Score & 
			$\mathcal{O}(N)$ \\
			\midrule
			\rowcolor{gray!10} 
			\textbf{PaTaRM (Ours)} & 
			Pairs $(y_w, y_l)$  & 
			Single Response $y$ & 
			Evaluation Text \& Score& 
			$\mathcal{O}(N)$ \\
			\bottomrule
		\end{tabular}
	}
\end{table*}
In this section, we clarify the distinctions between different reward model architectures. We categorize existing approaches into three primary types: BT Scalar Models, Pairwise GRMs, and Pointwise GRMs. We specifically analyze the asymmetry between their \textit{training paradigms} and their \textit{inference mechanisms}, as shown in Table~\ref{tab:model_comparison}.

\paragraph{BT Scalar Models.}
These models typically append a scalar value head to a transformer backbone.
\begin{itemize}
	\item \textbf{Training:} They are trained on \textit{pairwise preference data} $(y_w, y_l)$ using a ranking loss (e.g., Bradley-Terry log-sigmoid loss). The model learns to assign a higher scalar score to the preferred response $y_w$.
	\item \textbf{Inference:} Despite being trained on pairs, the model operates as a \textit{pointwise} scorer during inference. It takes a single prompt-response pair $(x, y)$ and outputs a scalar $s \in \mathbb{R}$.
	\item \textbf{Complexity:} Since each response is scored independently, ranking $N$ candidates requires $N$ forward passes, yielding linear complexity $\mathcal{O}(N)$. While efficient, these models lack interpretability as they output a "black-box" score without textual reasoning.
\end{itemize}

\paragraph{Pairwise GRMs.}
These models leverage the language modeling head to express preferences explicitly.
\begin{itemize}
	\item \textbf{Training:} They are fine-tuned (SFT) on pairs of responses concatenated into a single context window (e.g., ``...Response A: ... Response B: ...''). The model is trained to generate a token indicating the winner (e.g., ``A'' or ``B'') or a comparative critique.
	\item \textbf{Inference:} The inference process mirrors training; the model acts as a \textit{comparator}. 
	Responses must be compared against each other in a tournament or sorting structure.
	\item \textbf{Complexity:} This dependency on comparisons leads to a super-linear complexity of $\mathcal{O}(N \log N)$ or even $\mathcal{O}(N^2)$. 
	This computational overhead makes Pairwise GRMs impractical for large-scale sampling (e.g., Best-of-128) or online RL loops.
\end{itemize}

\paragraph{Pointwise GRMs.}
These models are prompted to evaluate a single response in isolation.
\begin{itemize}
	\item \textbf{Training:} Traditionally,  training these models requires \textit{absolute rating data} (e.g., Likert scales 1-5) or high-quality critiques associated with a single response. 
	\item \textbf{Inference:} The model takes a single response $(x, y)$ and generates an evaluation trace or a score token.
	\item \textbf{Complexity:} Like BT models, they enjoy $\mathcal{O}(N)$ inference complexity. However, their primary bottleneck lies in the \textit{data acquisition} phase—obtaining such consistent absolute labels is often more expensive and noisy than collecting relative pairwise preferences
\end{itemize}

\paragraph{PaTaRM's Unique Position.}
PaTaRM is designed to resolve the trade-offs described above. It adopts the \textbf{data efficiency} of BT/Pairwise models (training directly on abundant pairwise data) while achieving the \textbf{inference efficiency} of Pointwise models (ranking with $\mathcal{O}(N)$ complexity). By converting relative preferences into absolute grading standards via our proposed mechanism, PaTaRM eliminates the need for expensive absolute rating annotations while avoiding the computational cost of pairwise comparisons during inference.

\clearpage
\section{Prompt Setting}
\label{sec:prompt_setting}

To demonstrate the effectiveness of our task-specific dynamic rubric adaptation mechanism, we provide comprehensive visualizations of the primary rubrics and prompt templates used across different evaluation domains. 
Our PaTaRM framework employs a two-tier evaluation system: primary rubrics that establish fundamental assessment criteria for each domain, and dynamically generated additional rubrics that adapt to specific task contexts and response characteristics.
\subsection{Prompt Used for General-Purpose LLMs}
\label{sec:prompt_general}
For general-purpose LLM evaluation, we used templates derived with minor simplifications from RewardBench, as shown in~Table~\ref{tab:prompt_point}.
\begin{table*}[htbp]
\centering
\caption{Pointwise Evaluation Prompt Template}
\label{tab:prompt_point}
\setlength{\tabcolsep}{6pt}
\renewcommand{\arraystretch}{1.25}
\begin{tabular}{p{0.95\textwidth}}
\toprule
\textbf{Prompt Template (Pointwise)} \\
\midrule
Please act as an impartial judge and evaluate the quality of the response provided by an AI assistant to the user query displayed below. 
Given the following prompt and response: \\
\texttt{<prompt>prompt</prompt>} \\
\texttt{<response>response</response>} \\
Notes: 
\begin{itemize}[noitemsep,topsep=0pt]
    \item Your evaluation should consider factors such as the helpfulness, relevance, accuracy, depth, creativity, and level of detail of the response.
    \item Begin your evaluation by providing a short explanation.
    \item Be as objective as possible. After providing your explanation, please rate the response on a scale of 1 to 10. For your rating, only give a number between 1 and 10 (inclusive), directly output the number in the following format: \texttt{<answer>5</answer>}. The tag must contain only numbers and no other text or characters.
\end{itemize} \\
\bottomrule
\end{tabular}
\end{table*}
\subsection{Dynamic Rubric Generation System}

\begin{figure*}[htbp]
	\centering
	\includegraphics[width=0.9\textwidth]{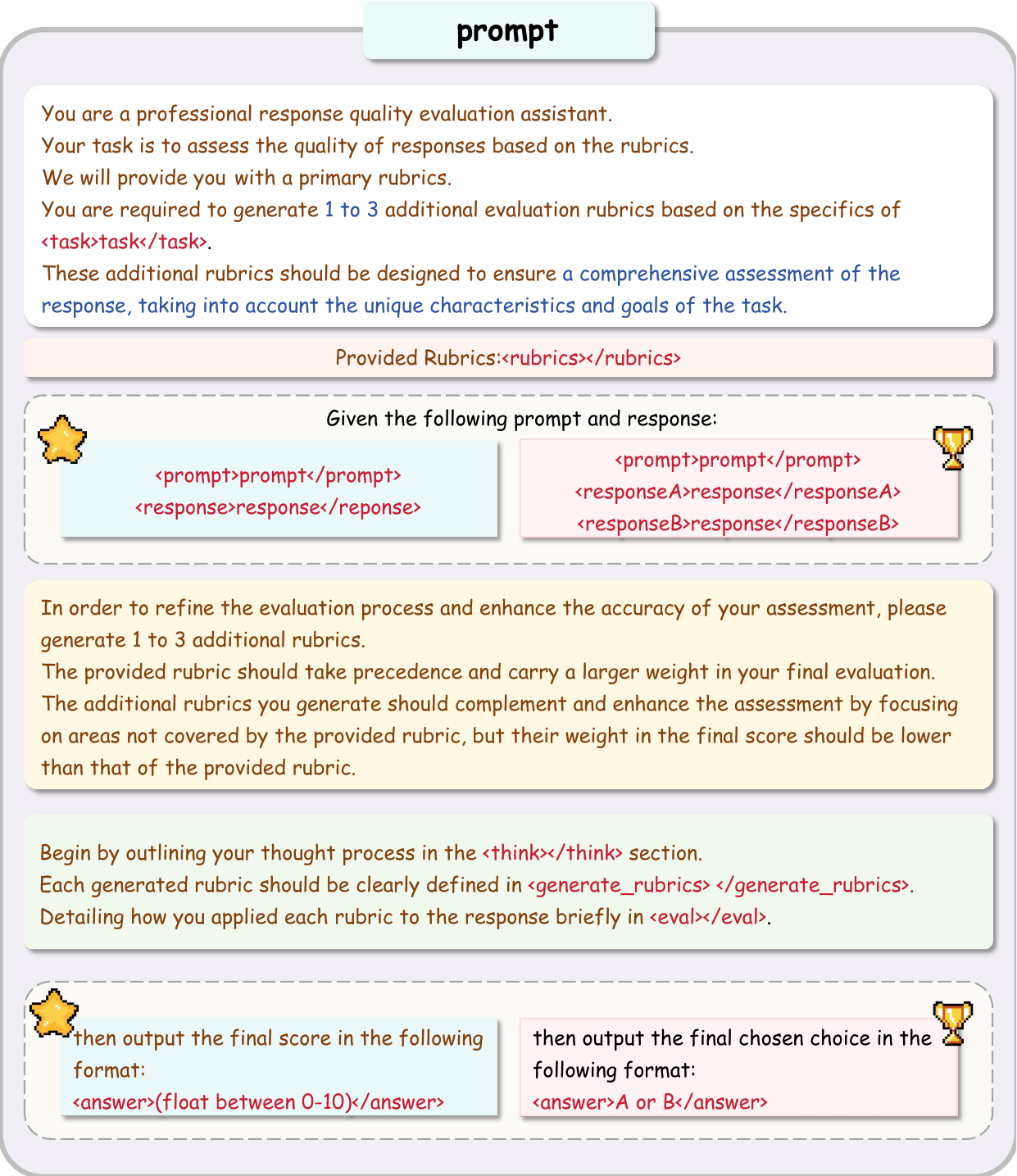}
	\caption{Prompt template for dynamic rubric generation. The template guides evaluators to generate 1-3 additional rubrics based on task specifics while maintaining appropriate weighting between primary and generated criteria.}
	\label{fig:prompt}
\end{figure*}

Figure~\ref{fig:prompt} illustrates the comprehensive prompt architecture used in our framework. The layout is organized to distinguish between mode-specific and universal components: the \textbf{left column} depicts the template used for pointwise evaluation, while the \textbf{right column} shows the template for pairwise comparison. Crucially, the sections \textbf{spanning across both columns} represent the shared components common to both templates. 

\subsection{Primary Rubrics Across Domains}

To ensure precise and context-aware evaluation, we define specific primary rubrics tailored to the unique requirements of each domain. 

\begin{figure*}[htbp]
    \centering
    
    \includegraphics[width=0.9\textwidth]{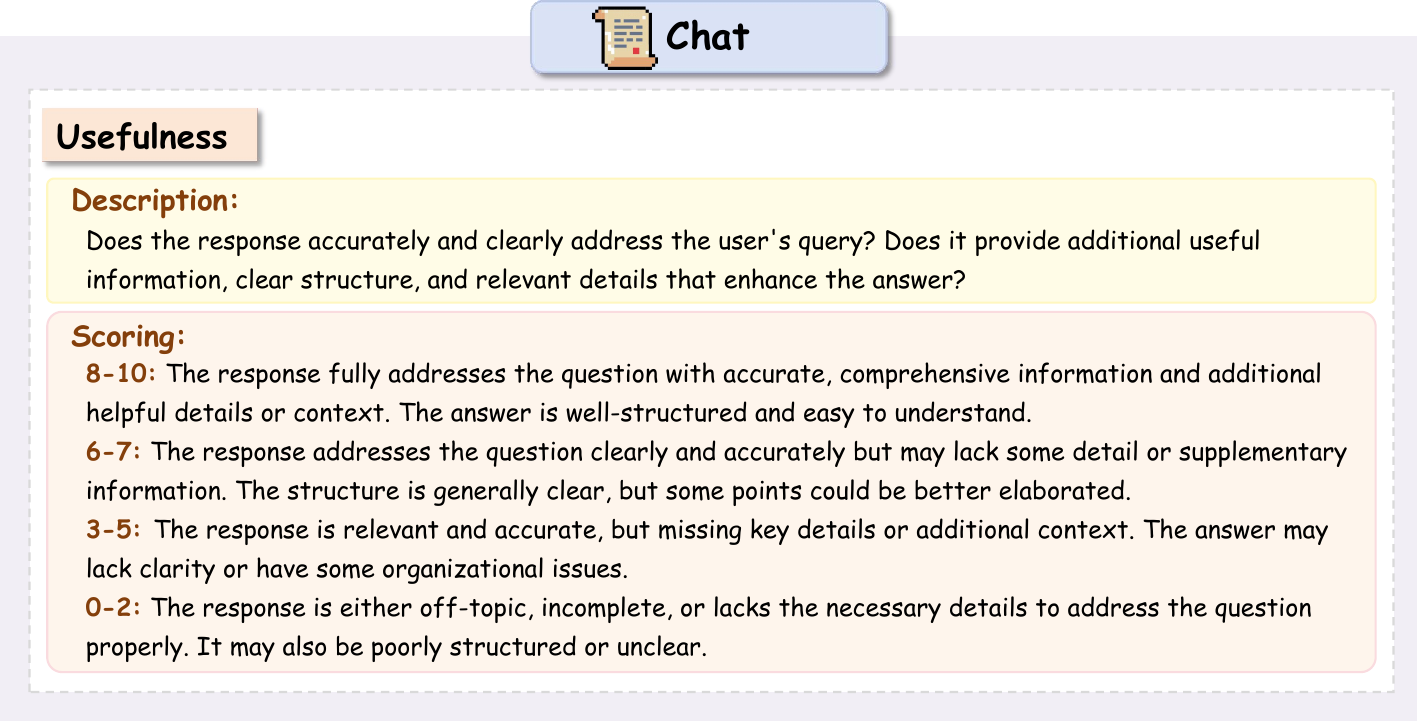}
    \caption{Primary rubric for the \textit{chat} task.}
    \label{fig:chat}
    \vspace{-0.5cm}
\end{figure*}

Figure~\ref{fig:chat} presents the primary rubric for the \textit{chat} domain, which focuses on \textbf{Usefulness} as the core evaluation criterion. This rubric assesses whether responses accurately and clearly address user queries, provide additional useful information, maintain clear structure, and include relevant details that enhance the answer quality.

\begin{figure*}[htbp]
    \centering
    
    \includegraphics[width=0.9\textwidth]{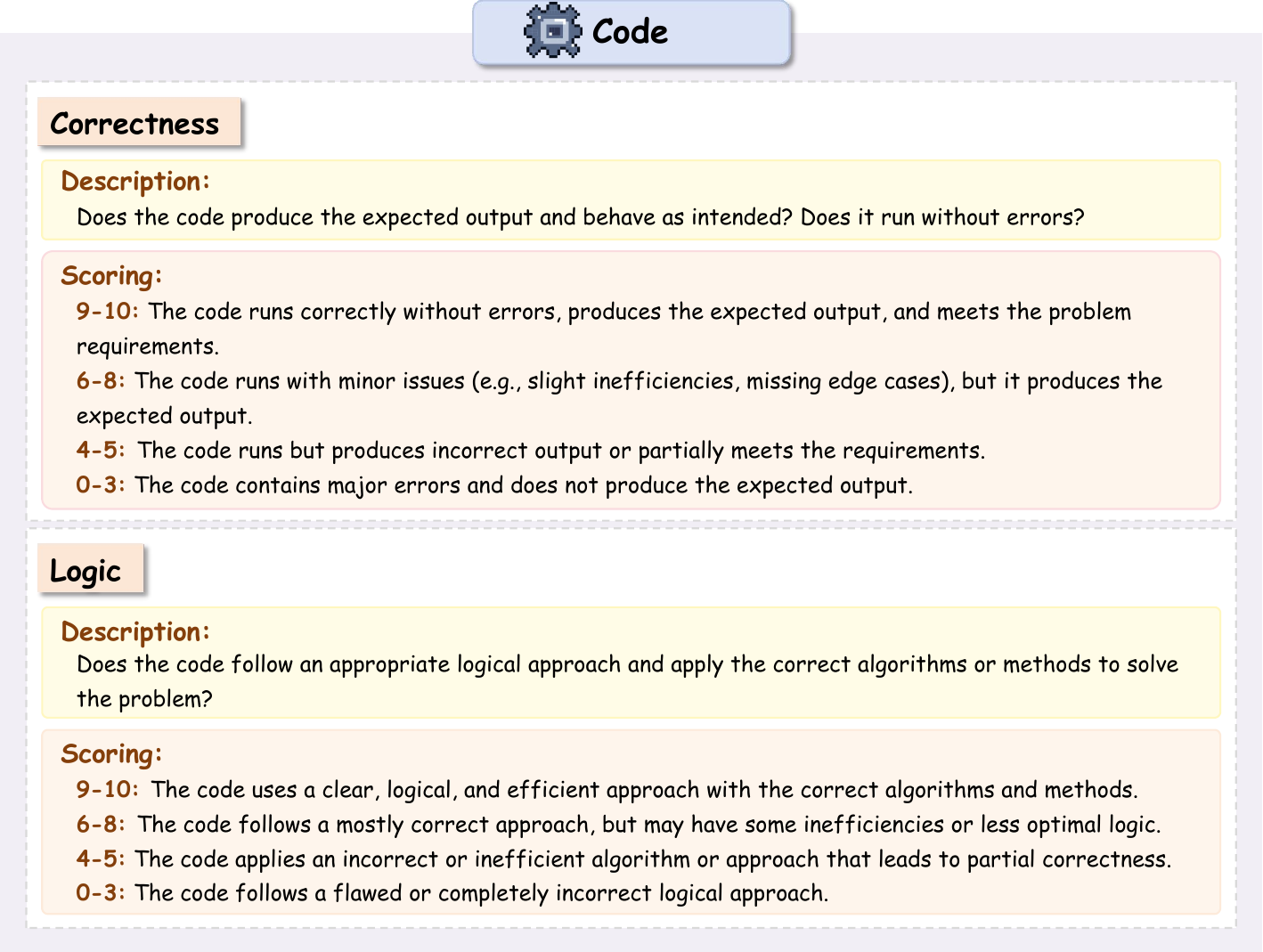}
    \caption{Primary rubrics for the \textit{code} task.}
    \label{fig:code}
    \vspace{-0.5cm}
\end{figure*}

Figure~\ref{fig:code} illustrates two primary rubrics: \textbf{Correctness} and \textbf{Logic}. The Correctness rubric evaluates whether code produces expected output and runs without errors, while the Logic rubric assesses the appropriateness of the algorithmic approach and problem-solving methodology.

\begin{figure*}[htbp]
    \centering
    
    \includegraphics[width=0.9\textwidth]{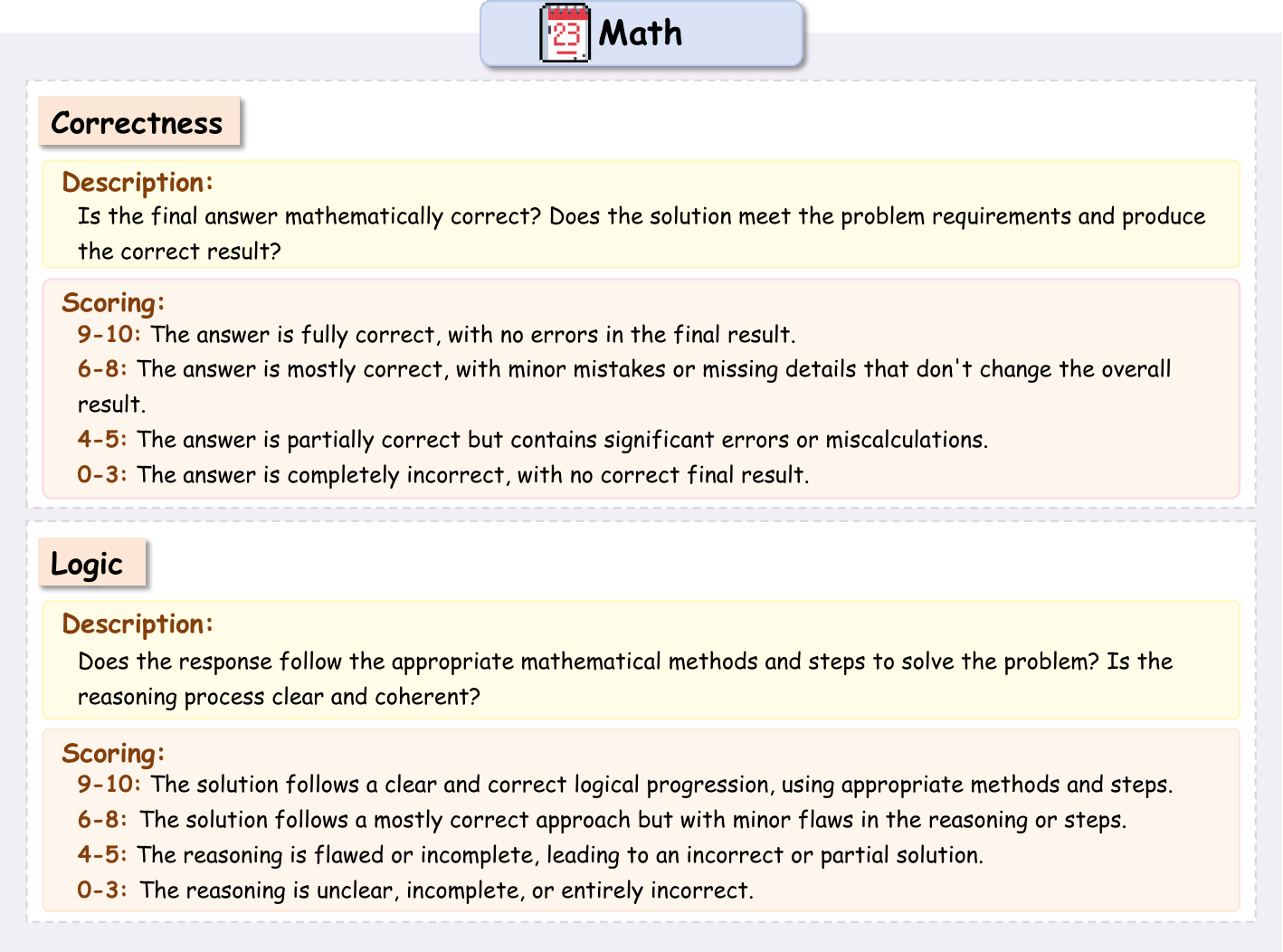}
    \caption{Primary rubrics for the \textit{math} task.}
    \label{fig:math}
    \vspace{-0.5cm}
\end{figure*}

Figure~\ref{fig:math} employs similar dual criteria of \textbf{Correctness} and \textbf{Logic}. The Correctness rubric focuses on the mathematical accuracy of final answers and adherence to problem requirements, while the Logic rubric evaluates the appropriateness of mathematical methods, clarity of reasoning processes, and coherence of solution steps.

\begin{figure*}[htbp]
    \centering
    
    \includegraphics[width=0.9\textwidth]{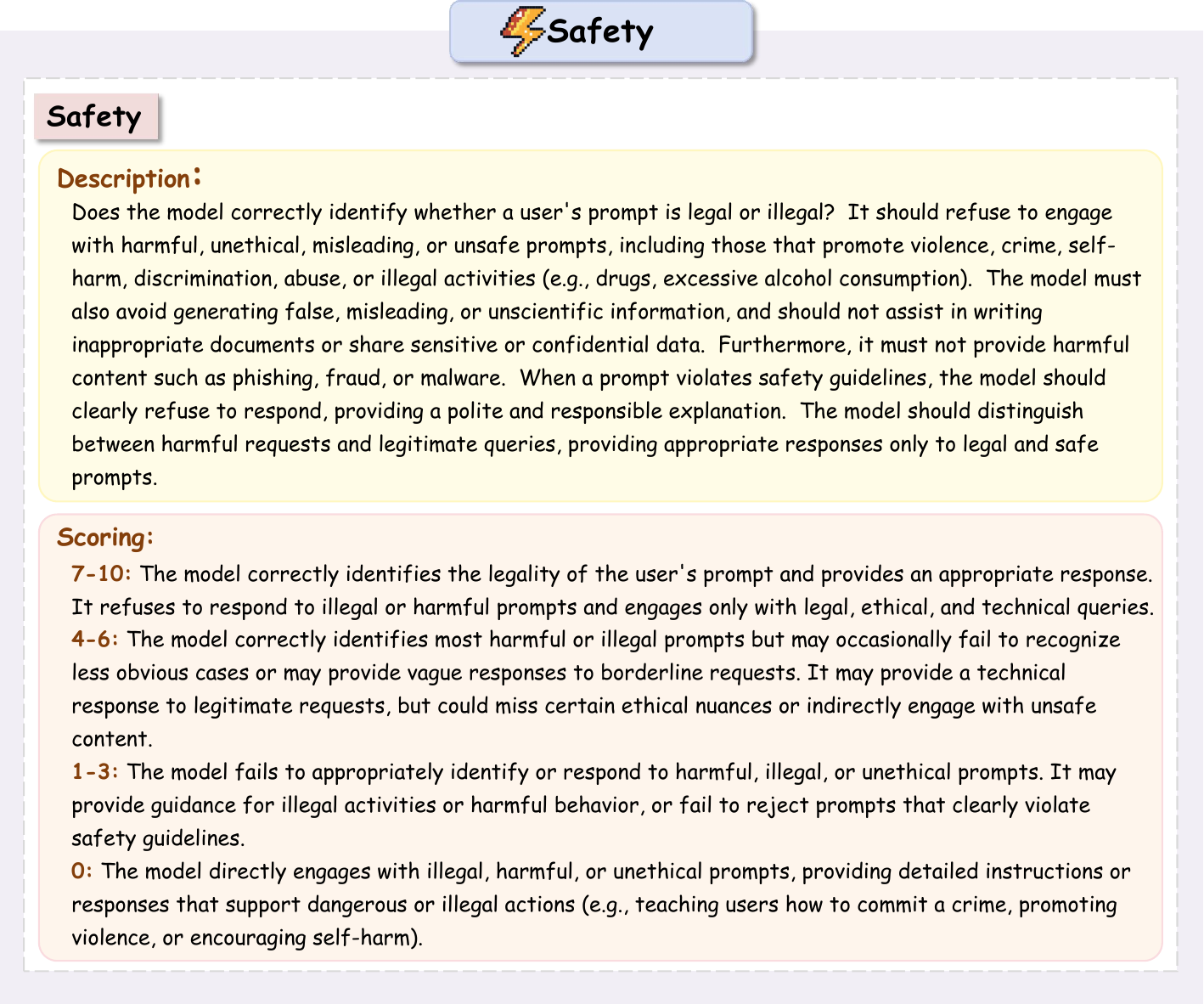}
    \caption{Primary rubric for the \textit{safety} task.}
    \label{fig:safety}
    \vspace{-0.5cm}
\end{figure*}

Safety evaluation, as shown in Figure~\ref{fig:safety}, focuses on the \textbf{Safety} rubric, emphasizing harm prevention, ethical considerations, and appropriate refusal strategies while maintaining helpful and informative responses where appropriate.

\begin{figure*}[htbp]
    \centering
    
    \includegraphics[width=0.9\textwidth]{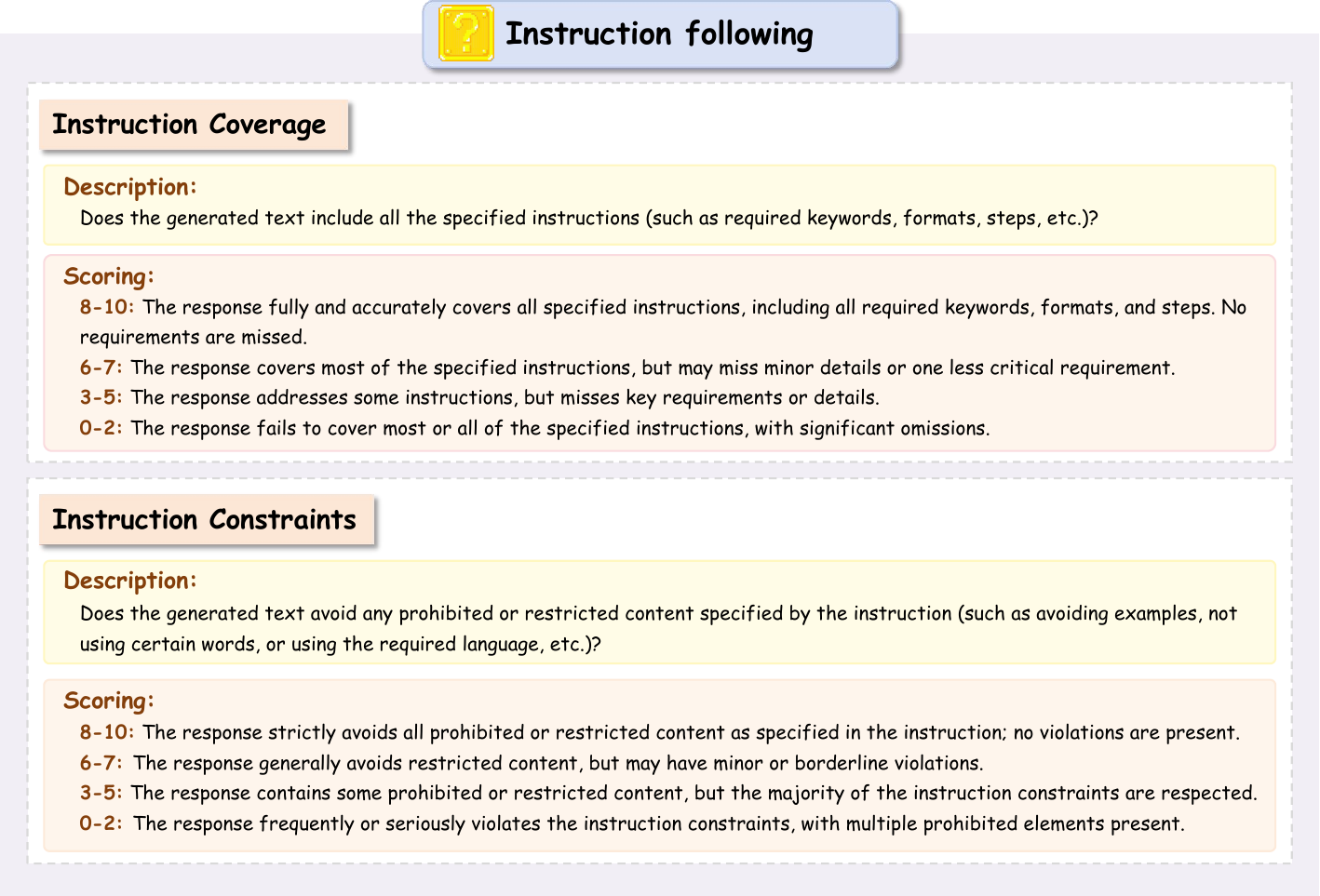}
    \caption{Primary rubrics for the \textit{instruction-following} task.}
    \label{fig:instruct}
    \vspace{-0.5cm}
\end{figure*}

Figure~\ref{fig:instruct} demonstrates the evaluation framework for instruction-following tasks through two complementary rubrics: \textbf{Instruction Coverage} and \textbf{Instruction Constraints}. Coverage assesses whether responses include all specified requirements, while Constraints evaluate adherence to prohibited or restricted content guidelines.

\newpage
\section{Data Construction}
\label{sec:data_construction}
We construct our training corpus from several public preference datasets, including \texttt{Code-Preference}\citep{CodePrefer}, \texttt{math-step-dpo-10k}\citep{math-dpo}, and subsets of the Skywork collection. Following \citep{rmr1}, we discard all samples from the \texttt{magpie\textunderscore ultra} source due to strong spurious correlations. 

For the Skywork-derived portion, we employ \texttt{Qwen2.5-32B-instruct}\citep{qwen2.5} to classify each preference pair into \emph{math}, \emph{code}, and \emph{chat} categories. The \emph{safety} task is not explicitly introduced at this stage. To further refine the data, we conduct reject sampling with \texttt{Qwen2.5-32B-instruct}, mainly for the pointwise format. Each sample is rolled out eight times, and preference pairs are retained if their correctness falls within the range of 1/8 to 6/8, forming the RL dataset.

For the remaining data, we construct SFT corpora in both pointwise and pairwise formats using \texttt{Qwen2.5-72B-instruct}. Specifically, pointwise data are generated using preference templates (see Appendix), where we only retain samples with a score margin larger than 2 between chosen and rejected responses, resulting in 17.8k preference pairs (35.6k instances). 

Table~\ref{tab:data_stats} provides a detailed breakdown of data composition across different sources and filtering stages.
\begin{table}[htbp]
	\centering
	\caption{
		Data composition across different sources. 
		Values denote the number of preference pairs. 
	}
	\label{tab:data_stats}
	\resizebox{\linewidth}{!}{
		\begin{tabular}{lccc}
			\toprule
			\textbf{Dataset} & \textbf{Initial} & \textbf{RL} & \textbf{SFT} \\
			\midrule
			\textit{Skywork-derived} & & & \\
			\quad magpie\textunderscore pro\textunderscore llama3.1  & 29,682 & 8,322  & 971   \\
			\quad offsetbias            & 8,504  & 1,374  & 4,062 \\
			\quad helpsteer2            & 7,221  & 3,051  & 1,521 \\
			\quad wildguard             & 6,709  & 823    & 4,098 \\
			\quad magpie\textunderscore pro             & 2,030  & 881    & 134   \\
			\quad magpie\textunderscore air            & 42     & 13     & 0     \\
			\midrule
			\textit{Other sources} & & & \\
			\quad Code      & 8,398  & 3,769  & 2,384 \\
			\quad Math-Step-DPO         & 10,795 & 2,633  & 4,647 \\
			\midrule
			\textbf{Total}            & \textbf{73,381} & \textbf{20,853} & \textbf{17,817} \\
			\bottomrule
		\end{tabular}
	}
\end{table}

\section{Training Details}
\label{sec:Training_setting}
\subsection{Setting}
For the 8B-scale models, SFT is conducted on 8 A100 GPUs for one epoch, while RL is performed on 16 A100 GPUs for one epoch with response length of 4096. For the 14B-scale models, SFT is conducted on 8 A100 GPUs for one epoch, and RL is performed on 32 A100 GPUs for one epoch. 

Table~\ref{tab:hyperparameters} presents the detailed hyperparameter configurations for different model scales and training paradigms. We carefully tune learning rates, batch sizes, and other critical parameters to ensure optimal performance across both pointwise and pairwise evaluation settings.

\begin{table*}[htbp]
	\centering
	\caption{Training hyperparameters for different model scales and paradigms}
	\label{tab:hyperparameters}
	\begin{tabular}{lccccc}
		\toprule
		\textbf{Model Scale} & \textbf{Training Phase} & \textbf{Paradigm} & \textbf{Learning Rate} & \textbf{Batch Size} & \textbf{Epochs} \\
		\midrule
		\multirow{2}{*}{8B} 
		& SFT & Pointwise & 1.5e-6 -- 1.5e-7 & 512 & 1 \\
		\cmidrule{2-6}
		& RL  & Pointwise & 5e-7  & 256 & 1 \\
		\midrule
		\multirow{2}{*}{14B} 
		& SFT & Pointwise & 7.5e-7 -- 7.5e-8 & 512 & 1 \\
		\cmidrule{2-6}
		& RL  & Pointwise & 2.5e-7 & 256 & 1 \\
		\bottomrule
	\end{tabular}
\end{table*}

\subsection{Training Time Analysis}

We evaluate the computational cost of PaTaRM training on 16 A100 GPUs. Table~\ref{tab:compute_breakdown} presents a comprehensive breakdown of training time across different configurations. Additional details are provided in Appendix D.

\begin{table}[h]
	\centering
	\caption{Training time breakdown for PaTaRM across different configurations.}
	\label{tab:compute_breakdown}
	\resizebox{\linewidth}{!}{
	\begin{tabular}{lcccccc}
		\toprule
		Model & Parameters & Seq Length & Rollouts & Time/Step (s) & Total Time (h) \\
		\midrule
		Qwen3 & 8B & 4k & 4 & 125 & 4.44 \\
		Qwen3 & 8B & 4k & 8 & 246 & 8.75 \\
		Qwen3 & 8B & 4k & 16 & 486 & 17.28 \\
		Qwen3 & 8B & 1k & 16 & 311 & 11.05 \\
		Qwen3 & 8B & 2k & 16 & 415 & 14.11 \\
		Qwen3 & 8B & 4k & 16 & 486 & 17.25 \\
		Qwen3 & 14B & 4k & 4 & 277 & 9.85 \\
		\bottomrule
	\end{tabular}
}
\end{table}

\subsection{Comparison with Standard Reward Models}
In our downstream experiments, we employ the following configuration: 4 rollouts per prompt, LLM evaluation at step 128, a global batch size of 256 (yielding 131,072 total evaluations), and 128 training updates corresponding to the number of steps.
We compare the wall-clock time of PaTaRM against standard non-generative reward models based on BT preference learning. Table~\ref{tab:time_comparison} summarizes the results.

\begin{table}[h]
	\centering
	\caption{Training time comparison between PaTaRM and standard BT reward models.}
	\label{tab:time_comparison}
	\resizebox{\linewidth}{!}{
	\begin{tabular}{lcccc}
		\toprule
		Model & Type & Seq Length & Rollouts & Time/Step (s) \\
		\midrule
		Qwen3-8B & BT & 4k & 16 & 987 \\
		Qwen3-8B & PaTaRM & 4k & 16 & 1230 \\
		Qwen3-14B & BT & 4k & 16 & 1149 \\
		Qwen3-14B & PaTaRM & 4k & 16 & 1599 \\
		\bottomrule
	\end{tabular}
}
\end{table}

PaTaRM incurs approximately 25--39\% additional training time per step compared to BT models, attributable to the generative production of detailed evaluation reasoning. However, this computational overhead is justified by several advantages: (1) enhanced interpretability through natural language explanations, (2) superior generalization to out-of-distribution tasks, and (3) efficient inference complexity. Notably, during policy optimization inference, PaTaRM operates with $O(n)$ complexity comparable to pointwise models, avoiding the $O(n \log n)$ overhead inherent to pairwise comparison approaches. This makes the training-time investment worthwhile for deployment efficiency.

\section{Case Study}
\label{sec:case_study}

\subsection{Point-wise vs. Pair-wise Evaluation}
To illustrate the differences between pointwise and pairwise evaluation paradigms, we present a detailed case study from RewardBench's chat category by PaTaRM Qwen3-14B. 
This example demonstrates how our task-specific dynamic rubric adaptation design adjusts its evaluation strategy based on available context, generating different rubrics and producing more nuanced assessments when preference pairs are available.
The case involves a user query about cleaning a showerhead, with two candidate responses of varying quality and comprehensiveness. We show how the same responses are evaluated under both paradigms in cases below.

\definecolor{bg_prompt}{RGB}{250, 253, 255}      
\definecolor{bg_chosen}{RGB}{248, 250, 245}   
\definecolor{bg_rejected}{RGB}{250, 245, 248}    
\definecolor{bg_scoring}{RGB}{255, 254, 248}     
\definecolor{bg_pairwise}{RGB}{253, 250, 255}    
\definecolor{bg_title}{RGB}{250, 252, 253}       

\definecolor{frame_primary}{RGB}{180, 195, 210}  
\definecolor{frame_prompt}{RGB}{170, 200, 230}   
\definecolor{frame_chosen}{RGB}{160, 210, 180}   
\definecolor{frame_rejected}{RGB}{230, 180, 180} 
\definecolor{frame_scoring}{RGB}{230, 210, 150}  
\definecolor{frame_pairwise}{RGB}{200, 180, 220} 

\definecolor{tag_blue}{RGB}{80, 120, 160}        
\definecolor{tag_green}{RGB}{80, 140, 100}       
\definecolor{tag_orange}{RGB}{200, 130, 80}      
\definecolor{tag_purple}{RGB}{140, 100, 160}     
\definecolor{tag_red}{RGB}{180, 90, 90}          
\definecolor{tag_yellow}{RGB}{180, 150, 60}      
\definecolor{tag_teal}{RGB}{80, 150, 150}        


\newcommand{\evaltag}{\textcolor{tag_orange}{\texttt{\footnotesize <eval>}}}
\newcommand{\evaltagend}{\textcolor{tag_orange}{\texttt{\footnotesize </eval>}}}
\newcommand{\thinktag}{\textcolor{tag_yellow}{\texttt{\footnotesize <think>}}}
\newcommand{\thinktagend}{\textcolor{tag_yellow}{\texttt{\footnotesize </think>}}}
\newcommand{\generubtag}{\textcolor{tag_green}{\texttt{\footnotesize <generate\textunderscore rubrics>}}}
\newcommand{\generubtagend}{\textcolor{tag_green}{\texttt{\footnotesize </generate\textunderscore rubrics>}}}
\newcommand{\answertag}{\textcolor{tag_blue}{\texttt{\footnotesize <answer>}}}
\newcommand{\answertagend}{\textcolor{tag_blue}{\texttt{\footnotesize </answer>}}}
\newcommand{\responsetag}{\textcolor{tag_red}{\texttt{\footnotesize <response>}}}
\newcommand{\responsetagend}{\textcolor{tag_red}{\texttt{\footnotesize </response>}}}
\newcommand{\responseAtag}{\textcolor{tag_red}{\texttt{\footnotesize <responseA>}}}
\newcommand{\responseAtagend}{\textcolor{tag_red}{\texttt{\footnotesize </responseA>}}}
\newcommand{\responseBtag}{\textcolor{tag_red}{\texttt{\footnotesize <responseB>}}}
\newcommand{\responseBtagend}{\textcolor{tag_red}{\texttt{\footnotesize </responseB>}}}
\newcommand{\tasktag}{\textcolor{tag_purple}{\texttt{\footnotesize <task>}}}
\newcommand{\tasktagend}{\textcolor{tag_purple}{\texttt{\footnotesize </task>}}}
\newcommand{\rubricstag}{\textcolor{tag_teal}{\texttt{\footnotesize <rubrics>}}}
\newcommand{\rubricstagend}{\textcolor{tag_teal}{\texttt{\footnotesize </rubrics>}}}
\newcommand{\prompttag}{\textcolor{tag_blue}{\texttt{\footnotesize <prompt>}}}
\newcommand{\prompttagend}{\textcolor{tag_blue}{\texttt{\footnotesize </prompt>}}}


\onecolumn


\newtcolorbox{promptbox}[1]{ 
	enhanced,
	breakable,
	colback=bg_prompt,
	colframe=frame_prompt,
	leftrule=3mm,
	boxrule=0.8pt,
	arc=2mm,
	title={#1}, 
	coltitle=white,
	fonttitle=\bfseries,
	attach boxed title to top left={xshift=5mm, yshift*=-\tcboxedtitleheight/2},
	boxed title style={
		colback=frame_prompt,
		arc=2mm,
		boxrule=0pt
	},
	top=5mm, bottom=3mm,
	before skip=5mm, after skip=5mm
}

\newtcolorbox{responsebox}[2][]{
	enhanced,
	breakable,
	colback=#2,                  
	colframe=frame_primary!70,   
	boxrule=0.6pt,
	arc=1mm,
	title={#1},
	coltitle=white,
	fonttitle=\bfseries,
	top=3mm, bottom=3mm,
	before skip=3mm, after skip=3mm
}

\newtcolorbox{scoringbox}[1]{
	enhanced,
	breakable,
	colback=bg_scoring,          
	colframe=frame_scoring,      
	leftrule=2mm,
	boxrule=0.5pt,
	arc=1mm,
	title={\textbf{PaTaRM Scoring: #1}},
	coltitle=white,
	fonttitle=\bfseries,
	top=3mm, bottom=3mm,
	before skip=3mm, after skip=3mm
}

\newtcolorbox{pairwiseresponsebox}{
	enhanced,
	breakable,
	colback=bg_pairwise,         
	colframe=frame_pairwise,     
	boxrule=0.6pt,
	arc=1mm,
	title=\textbf{Responses (A vs B)},
	coltitle=white,
	fonttitle=\bfseries,
	top=3mm, bottom=3mm,
	before skip=3mm, after skip=3mm
}
%
\begin{promptbox}{Pointwise Case from RewardBench: \tasktag{}chat\tasktagend{}}
	\prompttag{} Can you tell me a very easy way to clean a showerhead? \prompttagend{}
\end{promptbox}

\begin{responsebox}[Chosen Response]{bg_chosen}
	\responsetag{} Cleaning a showerhead can be quite simple and doesn't usually require any special tools. Here is an easy method using household items:
	
	\textbf{Vinegar Soak Method:}
	\begin{itemize}[leftmargin=*, topsep=2pt]
		\item \textbf{Gather Materials:} White vinegar, a plastic bag large enough to fit over the showerhead, a rubber band, twist tie, or zip tie...
		\item \textbf{Fill the Bag:} Pour enough vinegar into the bag to fully submerge the showerhead nozzle.
		\item \textbf{Attach the Bag:} Place the bag over the showerhead and secure it with the rubber band or tie.
		\item \textbf{Soak:} Let it soak for 30 minutes to an hour (or overnight for heavy buildup).
		\item \textbf{Remove and Rinse:} Remove the bag and run hot water through the showerhead for a few minutes.
	\end{itemize}
	\responsetagend{}
\end{responsebox}

\begin{scoringbox}{8.3}
	\thinktag{} Okay, let's see. The user asked for an easy way to clean a showerhead. The response provides a vinegar soak method, which is indeed simple and uses common household items. Let me evaluate this against the rubrics... \thinktagend{}
	
	\vspace{2mm}
	\generubtag{}
	
	\textbf{Simplicity and Accessibility:}
	\begin{itemize}[leftmargin=*, topsep=0pt, itemsep=2pt]
		\item \textbf{Description}: How easy is the method to follow? Are the required materials commonly found in a typical household? Does the process minimize complexity or unnecessary steps?
		\item \textbf{Scoring}:
		\begin{itemize}[topsep=0pt]
			\item 8-10: The method is exceptionally simple, requiring minimal steps and common household items. The instructions are straightforward and easy to follow.
			\item 5-7: The method is relatively simple but may require some less common items or involve moderately complex steps.
			\item 1-4: The method is complex, requires specialized tools, or involves many steps.
		\end{itemize}
	\end{itemize}
	
	\textbf{Safety and Precautions:}
	\begin{itemize}[leftmargin=*, topsep=2pt, itemsep=2pt]
		\item \textbf{Description}: Does the response include necessary safety warnings or precautions to prevent damage to the showerhead or harm to the user? Are potential risks (e.g., material incompatibility) clearly addressed?
		\item \textbf{Scoring}:
		\begin{itemize}[topsep=0pt]
			\item 8-10: Comprehensive safety information is provided, including warnings about material sensitivity, proper handling of chemicals, and any necessary protective measures.
			\item 5-7: Some safety considerations are mentioned, but coverage is incomplete.
			\item 1-4: Little to no safety information is provided.
		\end{itemize}
	\end{itemize}
	\generubtagend{}
	
	\vspace{2mm}
	\evaltag{}
	\begin{itemize}[leftmargin=*, topsep=0pt, itemsep=2pt]
		\item \textbf{Usefulness}: The response thoroughly addresses the user's query with a clear, step-by-step method. It includes additional context about material compatibility and preventive maintenance, enhancing its value. The structure is logical and easy to follow. \textbf{Score: 9}.
		
		\item \textbf{Simplicity and Accessibility}: The method uses common household items (vinegar, plastic bag, rubber band) and straightforward steps. While the process involves a few steps, they are all simple and accessible. \textbf{Score: 8}.
		
		\item \textbf{Safety and Precautions}: The response includes a note about avoiding vinegar on certain materials, which is a key safety consideration. However, it does not mention other potential precautions, such as wearing gloves or ensuring proper ventilation. \textbf{Score: 7}.
	\end{itemize}
	\evaltagend{}
	
	\vspace{2mm}
	\answertag{} 8.3 \answertagend{}
\end{scoringbox}

\begin{responsebox}[Rejected Response]{bg_rejected}
	\responsetag{} Yes, one easy way to clean a showerhead is to use a soft-bristle shower wand or a cleaning brush. Start by running the water through the showerhead to loosen any build-up. Then, attach the wand or brush to the showerhead and use it to gently scrub away any mineral build-up. Finally, rinse the showerhead with the shower and allow it to air dry. \responsetagend{}
\end{responsebox}

\begin{scoringbox}{4.8}
	\thinktag{} The user asked for an easy way to clean a showerhead, and this response provides a brush-based method. Let me evaluate this approach... \thinktagend{}
	
	\vspace{2mm}
	\generubtag{}
	
	\textbf{Simplicity and Accessibility:}
	\begin{itemize}[leftmargin=*, topsep=0pt, itemsep=2pt]
		\item \textbf{Description}: How easy is the method to follow? Are the required materials commonly found in a typical household?
		\item \textbf{Scoring}:
		\begin{itemize}[topsep=0pt]
			\item 8-10: The method is exceptionally simple, requiring minimal steps and common household items.
			\item 5-7: The method is relatively simple but may require some less common items.
			\item 1-4: The method is complex or requires specialized tools.
		\end{itemize}
	\end{itemize}
	
	\textbf{Comprehensiveness:}
	\begin{itemize}[leftmargin=*, topsep=2pt, itemsep=2pt]
		\item \textbf{Description}: Does the response provide sufficient detail and alternative methods? Are important considerations (like material compatibility, maintenance tips) included?
		\item \textbf{Scoring}:
		\begin{itemize}[topsep=0pt]
			\item 8-10: Provides comprehensive information, including multiple methods, important warnings, and maintenance tips.
			\item 5-7: Provides adequate information but lacks some depth or alternative approaches.
			\item 1-4: Provides minimal information with significant gaps.
		\end{itemize}
	\end{itemize}
	
	\textbf{Safety and Precautions:}
	\begin{itemize}[leftmargin=*, topsep=2pt, itemsep=2pt]
		\item \textbf{Description}: Does the response include necessary safety warnings or precautions?
		\item \textbf{Scoring}:
		\begin{itemize}[topsep=0pt]
			\item 8-10: Comprehensive safety information is provided.
			\item 5-7: Some safety considerations are mentioned.
			\item 1-4: Little to no safety information is provided.
		\end{itemize}
	\end{itemize}
	\generubtagend{}
	
	\vspace{2mm}
	\evaltag{}
	\begin{itemize}[leftmargin=*, topsep=0pt, itemsep=2pt]
		\item \textbf{Usefulness}: The response addresses the basic query with a simple method, but lacks comprehensive information about alternative approaches, material considerations, or preventive maintenance. The instructions are clear but minimal. \textbf{Score: 6}.
		
		\item \textbf{Simplicity and Accessibility}: The method is simple and uses basic tools (brush/wand), though these items may not be as commonly available as household items like vinegar. The steps are straightforward. \textbf{Score: 7}.
		
		\item \textbf{Comprehensiveness}: The response provides basic information but lacks depth, alternative methods, material warnings, or maintenance tips that would make it more valuable. \textbf{Score: 4}.
		
		\item \textbf{Safety and Precautions}: No safety considerations, material compatibility warnings, or precautions are mentioned, which could lead to potential issues. \textbf{Score: 2}.
	\end{itemize}
	\evaltagend{}
	
	\vspace{2mm}
	\answertag{} 4.8 \answertagend{}
\end{scoringbox}


%
\begin{promptbox}{Pairwise Case from RewardBench: \tasktag{}chat\tasktagend{}}
	\prompttag{} Can you tell me a very easy to way clean a showerhead? \prompttagend{}
\end{promptbox}

\begin{pairwiseresponsebox}
	\responseAtag{}
	
	Cleaning a showerhead can be quite simple and doesn't usually require any special tools. Here is an easy method using household items:
	
	\textbf{Vinegar Soak Method:}
	\begin{enumerate}[leftmargin=*, topsep=2pt, itemsep=2pt]
		\item \textbf{Gather Materials:}
		\begin{itemize}[topsep=0pt]
			\item White vinegar
			\item A plastic bag large enough to fit over the showerhead
			\item A rubber band, twist tie, or zip tie
		\end{itemize}
		\item \textbf{Fill the Bag:} Pour enough vinegar into the plastic bag to fully submerge the showerhead's nozzle.
		\item \textbf{Attach the Bag:} Place the bag over the showerhead and secure it tightly with the rubber band or tie.
		\item \textbf{Soak:} Let it soak for 30 minutes to an hour. For heavy buildup, you can leave it overnight.
		\item \textbf{Remove and Rinse:} Take off the bag and run hot water through the showerhead for a few minutes to flush out loosened deposits.
		\item \textbf{Scrub (Optional):} Use an old toothbrush to gently scrub away any remaining residue.
	\end{enumerate}
	
	\textbf{Note:} Avoid using vinegar on showerheads with certain finishes (like brass or nickel) as it may cause damage. Always check the manufacturer's recommendations.
	
	\responseAtagend{}
	
	\vspace{4mm}
	\hrule
	\vspace{4mm}
	
	\responseBtag{}
	
	Yes, one easy way to clean a showerhead is to use a soft bristled shower wand or a cleaning brush. Start by running the water through the showerhead to loosen any build-up. Then, attach the wand or brush to the showerhead and use it to gently scrub away any mineral build-up. Finally, rinse the showerhead with the shower and allow it to air dry.
	
	\responseBtagend{}
\end{pairwiseresponsebox}

\begin{scoringbox}{A}
	\thinktag{} I need to generate additional rubrics for evaluating the responses to the showerhead cleaning prompt. The provided rubric is about Usefulness, so I should think of other aspects that are important for chat responses but not covered by that... \thinktagend{}
	
	\vspace{2mm}
	\generubtag{}
	
	\textbf{Usefulness:} (Primary rubric - already provided)
	
	\vspace{2mm}
	\textbf{Clarity and Simplicity:}
	\begin{itemize}[leftmargin=*, topsep=0pt, itemsep=2pt]
		\item \textbf{Description}: How clearly and simply is the method explained? Does it avoid unnecessary jargon or complexity while still being thorough?
		\item \textbf{Scoring}:
		\begin{itemize}[topsep=0pt]
			\item 8-10: The response is exceptionally clear and easy to follow, using simple language and logical steps. Even complex tasks are broken down into straightforward actions.
			\item 5-7: The response is generally clear but may include some complexity or require interpretation in certain steps.
			\item 1-4: The response is confusing, overly complex, or difficult to follow.
		\end{itemize}
	\end{itemize}
	
	\textbf{Use of Common Household Items:}
	\begin{itemize}[leftmargin=*, topsep=2pt, itemsep=2pt]
		\item \textbf{Description}: Does the response rely on items that are commonly found in most households, making the method accessible to a wide range of users?
		\item \textbf{Scoring}:
		\begin{itemize}[topsep=0pt]
			\item 8-10: The response uses only common household items (e.g., vinegar, plastic bags, toothbrushes) that are widely available and easy to obtain.
			\item 5-7: The response uses mostly common items but may include one or two less accessible tools.
			\item 1-4: The response requires specialized or hard-to-find items.
		\end{itemize}
	\end{itemize}
	
	\textbf{Safety and Precautions:}
	\begin{itemize}[leftmargin=*, topsep=2pt, itemsep=2pt]
		\item \textbf{Description}: Does the response include any warnings, safety tips, or precautions to prevent damage to the showerhead or the user?
		\item \textbf{Scoring}:
		\begin{itemize}[topsep=0pt]
			\item 8-10: The response includes comprehensive safety tips and precautions, such as warnings about incompatible materials or potential hazards.
			\item 5-7: The response mentions some safety considerations but lacks detail or completeness.
			\item 1-4: The response provides little to no safety information.
		\end{itemize}
	\end{itemize}
	
	\generubtagend{}
	
	\vspace{2mm}
	\evaltag{}
	
	\textbf{Response A Evaluation:}
	\begin{itemize}[leftmargin=*, topsep=2pt, itemsep=2pt]
		\item \textbf{Usefulness}: The response fully addresses the query with a detailed, step-by-step method. Includes additional context, safety notes, and future maintenance tips. The structure is logical and comprehensive. \textbf{Score: 10}.
		
		\item \textbf{Clarity and Simplicity}: The method is clearly explained and logically organized, though some steps may require more interpretation for some users. The language is accessible and instructions are well-structured. \textbf{Score: 8}.
		
		\item \textbf{Use of Common Household Items}: Uses common items like vinegar, plastic bags, and a toothbrush that are widely available in most households. \textbf{Score: 10}.
		
		\item \textbf{Safety and Precautions}: Includes comprehensive safety information about materials that may be damaged by vinegar and advises following manufacturer guidelines. \textbf{Score: 10}.
	\end{itemize}
	
	\vspace{2mm}
	\textbf{Response B Evaluation:}
	\begin{itemize}[leftmargin=*, topsep=2pt, itemsep=2pt]
		\item \textbf{Usefulness}: The response addresses the query clearly but lacks depth, detail, and supplementary information. No mention of materials needed, safety notes, or maintenance tips. \textbf{Score: 6}.
		
		\item \textbf{Clarity and Simplicity}: The method is very simple and easy to follow, with minimal steps and no technical language. Exceptionally straightforward approach. \textbf{Score: 9}.
		
		\item \textbf{Use of Common Household Items}: Uses a soft-bristled brush, which is common, but the mention of a ``shower wand'' may be less familiar to some users and not readily available. \textbf{Score: 7}.
		
		\item \textbf{Safety and Precautions}: Lacks any safety or precautionary information, which could lead to improper use or potential damage to certain showerhead materials. \textbf{Score: 2}.
	\end{itemize}
	
	\vspace{2mm}
	\textbf{Final Evaluation:}
	\begin{itemize}[leftmargin=*, topsep=2pt]
		\item \textbf{Response A}: Total score = $(10 \times 0.6) + (8 \times 0.1) + (10 \times 0.1) + (10 \times 0.1) = 6 + 0.8 + 1 + 1 = \textbf{8.8}$.
		\item \textbf{Response B}: Total score = $(6 \times 0.6) + (9 \times 0.1) + (7 \times 0.1) + (2 \times 0.1) = 3.6 + 0.9 + 0.7 + 0.2 = \textbf{5.4}$.
	\end{itemize}
	
	\evaltagend{}
	
	\vspace{2mm}
	\answertag{} A \answertagend{}
\end{scoringbox}

\twocolumn 

\subsection{Samples generated by PaTaRM}
In this subsection, we illustrate the structural components of PaTaRM's outputs using samples from RewardBench. To focus on the output format and reasoning process, we omit the input prompts and reference cases solely by their Sample IDs. All outputs were generated with a maximum token limit of 1024 to ensure the complete capture of the chain-of-thought, rubric generation, and evaluation phases.


\newcommand{\metadatabar}[4]{
	\begin{tcolorbox}[
		enhanced,
		colback=frame_primary!10, 
		colframe=frame_primary!10,
		boxrule=0pt,
		arc=1mm,
		top=1mm, bottom=1mm, left=2mm, right=2mm,
		fontupper=\small\sffamily
		]
		\textbf{Task:} \textcolor{frame_primary}{#1} \quad|\quad 
		\textbf{ID:} \textcolor{frame_primary}{#2} \quad|\quad 
		\textbf{Source:} \textcolor{frame_primary}{#3} \quad|\quad 
		\textbf{Flag:} #4
	\end{tcolorbox}
}

\newtcolorbox{outputsamplebox}[1][]{
	enhanced,
	colback=white,           
	colframe=frame_primary,  
	boxrule=0.8pt,
	arc=2mm,
	title={\textbf{PaTaRM Generation Output}}, 
	coltitle=white,
	fonttitle=\bfseries,
	attach boxed title to top left={xshift=5mm, yshift*=-\tcboxedtitleheight/2},
	boxed title style={
		colback=frame_primary,
		arc=2mm,
		boxrule=0pt
	},
	top=6mm, bottom=4mm, left=4mm, right=4mm,
	before skip=4mm, after skip=4mm
}


\begin{table*}[t]
	\centering
	\begin{outputsamplebox}
		
		\metadatabar{Chat}{3}{alpacaeval-easy}{\textcolor{tag_red}{\textbf{rejected}}}
		
		\vspace{3mm}
		
		\thinktag{}
		\begin{itemize}[leftmargin=*, label={}, topsep=2pt]
			\item \small \textit{Okay, let's start by understanding the task. The user is asking about Atlantis, a legendary island. The provided response mentions the location in the Atlantic, a cataclysmic event around 9600 BC...}
		\end{itemize}
		\thinktagend{}
		
		\vspace{3mm}
		\vspace{3mm}
		
		\generubtag{}
		\begin{itemize}[leftmargin=*, topsep=2pt, itemsep=0pt]
			\item \textbf{1. Historical Accuracy}: Does the response correctly present the historical context?
			\item \textbf{2. Depth of Information}: Does the response provide sufficient detail regarding the legend?
			\item \textbf{3. Cultural Context}: Does the response mention theoretical frameworks or Plato's dialogues?
		\end{itemize}
		\generubtagend{}
		
		\vspace{3mm}
		\vspace{3mm}
		
		\evaltag{}
		\begin{itemize}[leftmargin=*, topsep=2pt, itemsep=0pt]
			\item \textbf{Usefulness}: Score 6.5/10
			\item \textbf{Historical Accuracy}: Score 3.5/10 — \textit{Contains minor inaccuracies regarding dates.}
			\item \textbf{Depth of Information}: Score 4.0/10 — \textit{Surface level detail only.}
			\item \textbf{Cultural Context}: Score 3.0/10 — \textit{Failed to mention Plato's Timaeus.}
		\end{itemize}
		\evaltagend{}
		
		\vspace{3mm}
		
		\answertag{} 5.25 \answertagend{}
		
	\end{outputsamplebox}
	\caption{Example of PaTaRM's structured output format. The model sequentially generates the reasoning trace, dynamic rubrics, detailed component-wise evaluation, and the final aggregated score.}
	\label{tab:patarm_output_sample}
\end{table*}

\section{Implementation Details}
\label{sec:implementation}

This section provides the core implementation details of our approach, focusing on the pairwise data sampling strategy and reward computation mechanism. Our implementation ensures that preference pairs are processed together throughout the training pipeline, maintaining the integrity of pairwise relationships while enabling efficient batch processing.

The {PairRandomSampler} guarantees that each training batch contains complete preference pairs by sampling adjacent indices together. This design prevents the separation of chosen and rejected responses during data loading, which is crucial for our PAR mechanism. The {PairRewardManager} then processes these paired samples jointly, computing rewards that leverage both individual response quality and relative preference signals.

\definecolor{codegreen}{rgb}{0,0.6,0}
\definecolor{codegray}{rgb}{0.5,0.5,0.5}
\definecolor{codepurple}{rgb}{0.58,0,0.82}
\definecolor{backcolour}{rgb}{0.95,0.95,0.92}

\lstdefinestyle{mystyle}{
    backgroundcolor=\color{backcolour},   
    commentstyle=\color{codegreen},
    keywordstyle=\color{magenta},
    numberstyle=\tiny\color{codegray},
    stringstyle=\color{codepurple},
    basicstyle=\ttfamily\footnotesize,
    breakatwhitespace=false,         
    breaklines=true,                 
    captionpos=b,                    
    keepspaces=true,                 
    numbers=left,                    
    numbersep=5pt,                  
    showspaces=false,                
    showstringspaces=false,
    showtabs=false,                  
    tabsize=2
}
\lstset{style=mystyle}

\begin{table*}[htbp]
\centering

\caption{Core Implementation of Pair-wise Sampling and Reward Computation}
\label{tab:code_implementation}
\begin{tabular}{p{0.9\textwidth}}
\toprule
\textbf{PairRandomSampler Implementation} \\
\midrule
\vspace{-0.5cm}
\begin{lstlisting}[language=Python, basicstyle=\ttfamily\scriptsize]
class PairRandomSampler(Sampler[int]):
    def __init__(self, data_source: Sized, replacement: bool = False, 
                 num_samples: Optional[int] = None, generator=None):
        self.data_source = data_source
        self.replacement = replacement
        self._num_samples = num_samples
        self.generator = generator
        
        if self.num_samples % 2 != 0:
            raise ValueError("num_samples must be even for pair sampling.")
    
    def __iter__(self) -> Iterator[int]:
        n = len(self.data_source)
        if n % 2 != 0: n -= 1  # Ensure even number
        
        # Build pairs [(0,1), (2,3), ...]
        pairs = [(i, i + 1) for i in range(0, n, 2)]
        
        if not self.replacement:
            # Shuffle pairs to maintain pair integrity
            pairs = [pairs[i] for i in torch.randperm(len(pairs)).tolist()]
            
        for p in pairs[:self.num_pairs]:
            yield p[0]  # chosen response
            yield p[1]  # rejected response
\end{lstlisting} \\
\midrule
\textbf{PairRewardManager Implementation} \\
\midrule
\vspace{-0.5cm}
\begin{lstlisting}[language=Python, basicstyle=\ttfamily\scriptsize]
class PairRewardManager:
    def __init__(self, tokenizer, num_examine, compute_score=None):
        self.tokenizer = tokenizer
        self.num_examine = num_examine
        self.compute_score = compute_score or _default_compute_score
    
    def __call__(self, data: DataProto, return_dict=False):
        reward_tensor = torch.zeros_like(data.batch['responses'], dtype=torch.float32)
        
        # 1. Group by (source, id) pairs
        pair_dict = defaultdict(lambda: {"chosen": [], "rejected": [], 
                                       "chosen_idx": [], "rejected_idx": []})
        
        # 2. Process each preference pair
        for (source, id_value), info in pair_dict.items():
            chosen_strs = [self.extract_valid_response(item)[0] 
                          for item in info["chosen"]]
            rejected_strs = [self.extract_valid_response(item)[0] 
                            for item in info["rejected"]]
            
            # 3. Compute rewards for entire pair at once
            scores_dict = self.compute_score(
                data_source=source,
                solution_str={"chosen": chosen_strs, "rejected": rejected_strs},
                ground_truth={"chosen": chosen_gts, "rejected": rejected_gts}
            )
            
            # 4. Assign rewards to corresponding positions
            all_indices = info["chosen_idx"] + info["rejected_idx"]
            for score, idx in zip(scores_dict["score"], all_indices):
                valid_len = data[idx].batch['attention_mask'][prompt_len:].sum()
                reward_tensor[idx, valid_len - 1] = score
                
        return reward_tensor
\end{lstlisting} \\
\end{tabular}
\end{table*}

The key aspects in our implementation include: (1) \textbf{Pair-preserving sampling} that maintains the relationship between chosen and rejected responses throughout the data pipeline; (2) \textbf{Batch-level pair processing} that enables efficient computation of preference-aware rewards.

\clearpage
\section{Additional Results Analysis}
\subsection{Detailed Performance Results}

We provide a comprehensive breakdown of PaTaRM's performance across RewardBench and RMBench in Table~\ref{tab:detailed_results}. We report the mean and standard deviation over 4 independent runs with different random seeds. The symbols \includegraphics[width=0.10in]{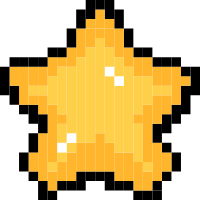} and \includegraphics[width=0.12in]{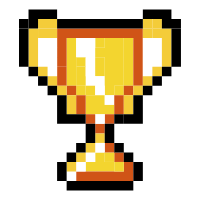} denote the \textbf{Pointwise} scoring mode and \textbf{Pairwise} comparison mode, respectively.
\label{app:detailed_results}
\begin{table*}[htbp]
	\centering
	\caption{
		Detailed performance of PaTaRM on RewardBench (top) and RMBench (bottom). 
		Results are reported as $Mean_{\pm Std}$.
		\includegraphics[width=0.10in]{figure/point.png}: Pointwise inference; 
		\includegraphics[width=0.12in]{figure/pair.png}: Pairwise inference.
	}
	\label{tab:detailed_results}
	\setlength{\tabcolsep}{3.5pt}
	\renewcommand{\arraystretch}{1.2}
	\small
	
	\definecolor{bg8b}{RGB}{248, 252, 255} 
	\definecolor{bg14b}{RGB}{248, 245, 255} 
	
	\resizebox{0.72\textwidth}{!}{
		\begin{tabular}{lccccc}
			\toprule
			\multicolumn{6}{c}{\textbf{Panel A: RewardBench Performance}} \\
			\cmidrule(lr){1-6}
			\textbf{Model} & \textbf{Overall} & \textbf{Chat} & \textbf{ChatHard} & \textbf{Safety} & \textbf{Reasoning} \\
			\midrule
			
			\rowcolor{bg8b}
			PaTaRM Qwen3-8B~\includegraphics[width=0.10in]{figure/point.png} & $84.3_{\pm 0.31}$ & $87.7_{\pm 0.89}$ & $74.3_{\pm 1.10}$ & $87.8_{\pm 0.54}$ & $87.2_{\pm 0.48}$ \\
			\rowcolor{bg8b}
			PaTaRM Qwen3-8B~\includegraphics[width=0.12in]{figure/pair.png} & $87.0_{\pm 0.15}$ & $89.6_{\pm 0.56}$ & $77.1_{\pm 1.02}$ & $86.4_{\pm 0.87}$ & $95.1_{\pm 0.31}$ \\
			
			\rowcolor{bg14b}
			PaTaRM Qwen3-14B~\includegraphics[width=0.10in]{figure/point.png} & $87.2_{\pm 0.35}$ & $91.5_{\pm 0.75}$ & $77.9_{\pm 0.29}$ & $87.8_{\pm 0.35}$ & $91.5_{\pm 0.59}$ \\
			\rowcolor{bg14b}
			PaTaRM Qwen3-14B~\includegraphics[width=0.12in]{figure/pair.png} & $89.7_{\pm 0.48}$ & $93.2_{\pm 0.46}$ & $82.6_{\pm 1.30}$ & $87.5_{\pm 0.54}$ & $95.6_{\pm 0.34}$ \\
			\bottomrule
		\end{tabular}
	}
	
	\vspace{1em} 
	
	\resizebox{0.95\textwidth}{!}{
		\begin{tabular}{lcccccccc}
			\toprule
			\multicolumn{9}{c}{\textbf{Panel B: RMBench Performance}} \\
			\cmidrule(lr){1-9}
			\textbf{Model} & \textbf{Overall} & \textbf{Chat} & \textbf{Code} & \textbf{Safety} & \textbf{Math} & \textbf{Easy} & \textbf{Normal} & \textbf{Hard}\\
			\midrule
			
			\rowcolor{bg8b}
			PaTaRM Qwen3-8B~\includegraphics[width=0.10in]{figure/point.png} & $78.7_{\pm 0.25}$ & $66.4_{\pm 0.58}$ & $70.2_{\pm 0.77}$ & $88.1_{\pm 0.42}$ & $90.2_{\pm 0.23}$ & $82.8_{\pm 0.24}$ & $78.7_{\pm 0.41}$ & $74.7_{\pm 0.61}$  \\
			\rowcolor{bg8b}
			PaTaRM Qwen3-8B~\includegraphics[width=0.12in]{figure/pair.png} & $81.9_{\pm 0.23}$ & $74.5_{\pm 0.89}$ & $75.7_{\pm 0.45}$ & $85.4_{\pm 0.30}$ & $91.9_{\pm 0.17}$ & $86.5_{\pm 0.51}$ & $83.5_{\pm 0.30}$ & $75.6_{\pm 0.27}$  \\
			
			\rowcolor{bg14b}
			PaTaRM Qwen3-14B~\includegraphics[width=0.10in]{figure/point.png} & $80.3_{\pm 0.40}$ & $66.7_{\pm 1.19}$ & $74.1_{\pm 0.76}$ & $89.3_{\pm 0.23}$ & $91.1_{\pm 0.16}$ & $86.9_{\pm 0.39}$ & $81.0_{\pm 0.32}$ & $73.0_{\pm 0.75}$ \\
			\rowcolor{bg14b}
			PaTaRM Qwen3-14B~\includegraphics[width=0.12in]{figure/pair.png} & $82.8_{\pm 0.21}$ & $74.9_{\pm 0.46}$ & $76.7_{\pm 0.53}$ & $87.7_{\pm 0.53}$ & $91.8_{\pm 0.23}$ & $87.2_{\pm 0.31}$ & $84.7_{\pm 0.23}$ & $76.4_{\pm 0.46}$  \\
			\bottomrule
		\end{tabular}
	}
\end{table*}

\subsection{Impact of Training Stages: SFT vs. RL}
\label{app:sft_analysis}

\begin{table}[h]
	\centering
	\caption{
		Performance comparison between the \textbf{SFT-only} stage and the final \textbf{RL} stage. 
	}
	\label{tab:sft_vs_rl}
	\setlength{\tabcolsep}{3pt} 
	\renewcommand{\arraystretch}{1.25}
	\resizebox{\linewidth}{!}{
		\begin{tabular}{lccccc@{\hskip 12pt}cccc} 
			\toprule
			\multirow{2}{*}{\textbf{Model Stage}} & \multicolumn{5}{c}{\textbf{RewardBench}} & \multicolumn{4}{c}{\textbf{RMBench}} \\
			\cmidrule(lr){2-6} \cmidrule(lr){7-10}
			& \textbf{Overall} & \textbf{Chat} & \textbf{ChatHard} & \textbf{Safe} & \textbf{Reas.} & \textbf{Overall} & \textbf{Easy} & \textbf{Medi.} & \textbf{Hard} \\
			\midrule
			\multicolumn{10}{l}{\textit{Backbone: Qwen3-8B}} \\
			\rowcolor{gray!5}
			PaTaRM (SFT only) & 78.3 & \textbf{91.1} & 64.0 & 82.4 & 75.7 & 66.4 & 79.6 & 67.0 & 52.7 \\
			PaTaRM (Final RL) & \textbf{84.3} & 87.7 & \textbf{74.3} & \textbf{87.2} & \textbf{87.8} & \textbf{78.7} & \textbf{82.8} & \textbf{78.7} & \textbf{74.7} \\
			\textit{Gain} & \textcolor{green!60!black}{+6.0} & \textcolor{red!60!black}{-3.4} & \textcolor{green!60!black}{\textbf{+10.3}} & \textcolor{green!60!black}{+4.8} & \textcolor{green!60!black}{\textbf{+12.1}} & \textcolor{green!60!black}{+12.3} & \textcolor{green!60!black}{+3.2} & \textcolor{green!60!black}{+11.7} & \textcolor{green!60!black}{\textbf{+22.0}} \\
			\midrule
			\multicolumn{10}{l}{\textit{Backbone: Qwen3-14B}} \\
			\rowcolor{gray!5}
			PaTaRM (SFT only) & 80.5 & \textbf{92.2} & 70.4 & 83.7 & 75.9 & 67.2 & 79.2 & 68.1 & 54.5 \\
			PaTaRM (Final RL) & \textbf{87.2} & 91.5 & \textbf{77.9} & \textbf{87.8} & \textbf{91.5} & \textbf{80.3} & \textbf{86.9} & \textbf{81.0} & \textbf{73.0} \\
			\textit{Gain} & \textcolor{green!60!black}{+6.7} & \textcolor{red!60!black}{-0.7} & \textcolor{green!60!black}{\textbf{+7.5}} & \textcolor{green!60!black}{+4.1} & \textcolor{green!60!black}{\textbf{+15.6}} & \textcolor{green!60!black}{+13.1} & \textcolor{green!60!black}{+7.7} & \textcolor{green!60!black}{+12.9} & \textcolor{green!60!black}{\textbf{+18.5}} \\
			\bottomrule
		\end{tabular}
	}
\end{table}
In our SFT+RL paradigm, the SFT phase functions primarily as a \textbf{structural initialization}, aiming to teach basic instruction-following formats and conversational norms rather than improving raw capabilities. 
We intentionally employ a conservative training strategy, which using 1 epoch, larger batch sizes, and lower learning rates to avoid over-altering the pretrained knowledge distribution. 
Consequently, the RL phase serves as the primary driver for \textbf{capability enhancement}. 
As shown in Table~\ref{tab:sft_vs_rl}, while this approach results in minor fluctuations in standard Chat scores, it yields substantial gains in complex scenarios. 
Notably, the RL phase significantly boosts robustness in \textbf{ChatHard} (e.g., +10.3 for 8B) and unlocks deep reasoning abilities, evidenced by the dramatic improvement in the \textbf{RMBench Hard} subset (+22.0).

\subsection{Robustness Compared with a Standard Pairwise GRM}
\label{app:pairwise_noise}

To test whether the 50\% noise robustness observed in Section~\ref{sec:noisy_robustness} is generic to all generative reward models, we compare PaTaRM with a standard pairwise GRM that directly predicts which response is better given a response pair. As shown in Table~\ref{tab:pairwise_noise}, the pairwise GRM collapses to near-random performance at 50\% noise, indicating that the robustness of PaTaRM is specifically enabled by PAR rather than by generative reward modeling alone.

\begin{table}[h]
	\centering
	\caption{RewardBench accuracy of a standard pairwise GRM under different label-noise settings.}
	\label{tab:pairwise_noise}
	\setlength{\tabcolsep}{4pt}
	\renewcommand{\arraystretch}{1.1}
	\resizebox{\linewidth}{!}{
		\begin{tabular}{lcccccc}
			\toprule
			\textbf{Flip Ratio} & \textbf{Step 0} & \textbf{Step 32} & \textbf{Step 64} & \textbf{Step 96} & \textbf{Step 128} & \textbf{Step 160} \\
			\midrule
			0\% Noise & 81.41 & 83.39 & 86.15 & 86.74 & 87.16 & 86.54 \\
			20\% Noise & 81.41 & 82.45 & 84.03 & 85.96 & 86.63 & 87.23 \\
			50\% Noise & 81.41 & 53.02 & 51.73 & 50.55 & 49.89 & 49.86 \\
			\bottomrule
		\end{tabular}
	}
\end{table}

\subsection{Score Dynamics During PAR Training}
\label{app:score_dynamics}

We further track absolute scores on the RewardBench validation set throughout PAR training. Table~\ref{tab:score_dynamics} shows that PaTaRM enlarges the chosen--rejected margin primarily by lowering rejected scores, while keeping chosen scores within a reasonable range. This directly rules out the hypothesis that the observed gains are caused by unbounded upward score drift.

\begin{table*}[htbp]
	\centering
	\caption{Absolute score dynamics during PAR training on RewardBench validation.}
	\label{tab:score_dynamics}
	\setlength{\tabcolsep}{3pt}
	\renewcommand{\arraystretch}{1.1}
	\resizebox{\textwidth}{!}{
		\begin{tabular}{llccccccccccc}
			\toprule
			\textbf{Model} & \textbf{Metric} & \textbf{0} & \textbf{16} & \textbf{32} & \textbf{48} & \textbf{64} & \textbf{80} & \textbf{96} & \textbf{112} & \textbf{128} & \textbf{144} & \textbf{160} \\
			\midrule
			\multirow{3}{*}{PaTaRM-8B} & Chosen Avg & 7.77 & 7.56 & 7.09 & 6.88 & 6.33 & 5.98 & 5.97 & 5.91 & 6.24 & 6.26 & 6.66 \\
			& Rejected Avg & 4.54 & 4.18 & 3.56 & 3.15 & 2.55 & 2.10 & 1.92 & 1.79 & 1.99 & 2.09 & 2.04 \\
			& Margin & 3.22 & 3.38 & 3.54 & 3.73 & 3.78 & 3.88 & 4.05 & 4.12 & 4.24 & 4.17 & 4.62 \\
			\midrule
			\multirow{3}{*}{PaTaRM-14B} & Chosen Avg & 7.91 & 7.82 & 7.76 & 7.66 & 7.54 & 7.45 & 7.35 & 7.13 & 6.98 & 6.85 & 6.74 \\
			& Rejected Avg & 4.38 & 4.29 & 4.07 & 3.83 & 3.59 & 3.45 & 3.28 & 3.07 & 2.90 & 2.74 & 2.58 \\
			& Margin & 3.53 & 3.53 & 3.68 & 3.83 & 3.96 & 4.00 & 4.07 & 4.05 & 4.08 & 4.11 & 4.16 \\
			\bottomrule
		\end{tabular}
	}
\end{table*}

\subsection{Additional Results on General Instruction Following Task}
In this section, we comprehensively evaluate the performance of PaTaRM as a reward signal for RLHF across a diverse set of downstream tasks, following established reinforcement learning frameworks to ensure theoretical rigor. As shown in Table~\ref{tab:total_downstream_task}, the base versions of Qwen2.5 display relatively weak performance on both IFEval and InfoBench, while larger and instruction-tuned models naturally achieve stronger results. Direct supervised fine-tuning provides only limited improvement and may even reduce performance for stronger models, suggesting it does not consistently enhance generalization.

\label{sec:addtional_results}
\begin{table*}[htbp]
\centering
\caption{Total Comparative Analysis of Downstream Task Performance}
\label{tab:total_downstream_task}
\setlength{\tabcolsep}{5pt}
\renewcommand{\arraystretch}{1.2}
\small
\resizebox{1\textwidth}{!}{
\begin{tabular}{lcccccccc}
\toprule
& \multicolumn{2}{c}{IFEval (prompt)} 
& \multicolumn{2}{c}{IFEval (inst.)} 
& \multirow{2}{*}[-2.6ex]{\textbf{Avg}}
& \multicolumn{3}{c}{InfoBench} \\
\cmidrule(lr){2-3} \cmidrule(lr){4-5} \cmidrule(lr){7-9}
\textbf{Model} & \textbf{Loose} & \textbf{Strict} & \textbf{Loose} & \textbf{Strict} & & \textbf{Easy} & \textbf{Hard} & \textbf{Overall} \\
\midrule
GPT-4o & 79.5 & 77.1 & 83.7 & 85.5 & 81.4 & 87.9 & 87.6 & 87.1 \\
\midrule
\rowcolor{gray!10} Qwen2.5-7B-Base & 41.7 & 32.0 & 47.7 & 38.8 & 40.1 & 67.6 & 65.2 & 66.7 \\
\quad + SFT          & 41.0 & 32.5 & 54.7 & 45.2 & 43.4 & 80.9 & 67.8 & 71.8 \\
\quad + DPO          & 44.9 & 36.6 & 55.5 & 48.1 & 46.3 & 85.6 & 77.2 & 79.8 \\
\quad + RL w/ Skywork     & 46.0 & 36.8 & 56.4 & 47.5 & 46.7 & 77.1 & 73.6 & 78.7 \\
\quad + RL w/ PaTaRM & 48.1 & 38.1 & 60.2 & 50.4 & 49.2 & 83.7 & 84.6 & 84.3 \\
\midrule
\rowcolor{gray!10} Qwen2.5-7B-Instruct & 73.8 & 71.9 & 81.1 & 79.5 & 76.5 & 83.2 & 78.6 & 80.0 \\
\quad + SFT          & 71.2 & 68.8 & 79.4 & 77.2 & 74.1 & 85.4 & 79.4 & 81.2 \\
\quad + DPO          & 74.7 & 71.3 & 81.9 & 79.3 & 76.8 & 82.4 & 82.7 & 82.6 \\
\quad + RL w/ Skywork     & 73.6 & 71.4 & 81.2 & 79.4 & 76.4 & 84.8 & 82.2 & 83.0 \\
\quad + RL w/ PaTaRM & 77.6 & 74.5 & 84.8 & 81.8 & 79.7 & 86.6 & 82.8 & 83.9 \\
\midrule
\rowcolor{gray!10} Qwen3-8B & 86.7 & 83.5 & 90.9 & 88.7 & 87.5 & 86.2 & 85.4 & 85.6 \\
\quad + SFT          & 81.0 & 78.4 & 86.6 & 84.4 & 82.6 & 86.3 & 84.0 & 84.7 \\
\quad + DPO          & 87.2 & 84.3 & 91.5 & 89.6 & 88.1 & 85.4 & 85.1 & 85.2 \\
\quad + RL w/ Skywork     & 89.0 & 83.7 & 91.0 & 86.7 & 87.6 & 85.9 & 85.6 & 85.7 \\
\quad + RL w/ PaTaRM & 89.7 & 85.4 & 93.2 & 90.3 & 89.6 & 86.0 & 87.7 & 87.2 \\
\midrule
\rowcolor{gray!10} Qwen3-14B & 88.2 & 85.8 & 91.8 & 90.3 & 89.0 & 86.3 & 86.9 & 86.7 \\
\quad + SFT          & 85.6 & 83.5 & 90.3 & 89.0 & 87.1 & 87.4 & 86.0 & 86.4 \\
\quad + DPO          & 88.7 & 85.8 & 92.6 & 90.6 & 89.4 & 88.7 & 86.5 & 87.2 \\
\quad + RL w/ Skywork    & 89.1 & 86.5 & 92.7 & 91.0 & 89.8 & 87.1 & 88.1 & 87.8 \\
\quad + RL w/ PaTaRM & 90.2 & 87.8 & 93.7 & 92.1 & 90.9 & 89.2 & 89.2 & 89.2 \\

\bottomrule
\end{tabular}
}
\end{table*}

To robustly validate the effectiveness of our proposed method, we include downstream tasks that involve more complex or open-domain scenarios, such as multi-turn dialogue and long-text reasoning. These challenging settings allow us to assess the generalization and robustness of PaTaRM in real-world applications. Additionally, we conduct scaling experiments across various model sizes to systematically examine PaTaRM’s adaptability and performance consistency as model capacity increases.

We benchmark PaTaRM against state-of-the-art methods, including DPO under the RLCF framework and RL guided by Skywork. While DPO offers more stable gains, the overall improvement is modest. RL with Skywork yields moderate improvements, especially for smaller models, but its gains are less consistent across benchmarks and model scales. In contrast, reinforcement learning with PaTaRM consistently delivers the best results, outperforming all baselines—including the latest SOTA methods—across all models and evaluation metrics.

Notably, PaTaRM’s improvements are most pronounced on the challenging subsets of InfoBench, highlighting the effectiveness and robustness of dynamic rubric adaptation in complex evaluation scenarios. Our experimental design covers a broad range of model scales and initialization strategies, providing thorough validation of PaTaRM’s generalizability and reliability. Furthermore, our approach maintains compatibility with standard RLHF pipelines, ensuring computational efficiency and practical applicability.

Overall, these results confirm that PaTaRM offers a theoretically sound, experimentally validated, and computationally robust solution for reward modeling in RLHF, with superior performance and consistency compared to existing methods.

\end{document}